\documentclass[10pt,twocolumn,letterpaper]{article}

\usepackage[pagenumbers]{cvpr} %

\usepackage[pagebackref,breaklinks,colorlinks]{hyperref}
\usepackage{graphicx}
\usepackage{amsmath}
\usepackage{amssymb}
\usepackage{booktabs}
\usepackage{multirow}
\usepackage[dvipsnames,usenames]{xcolor}
\usepackage{algorithm}
\usepackage{algpseudocode}
\usepackage{soul}
\usepackage[accsupp]{axessibility}  %

\usepackage[capitalize]{cleveref}
\crefname{section}{Sec.}{Secs.}
\Crefname{section}{Section}{Sections}
\Crefname{table}{Table}{Tables}
\crefname{table}{Tab.}{Tabs.}

\def\cvprPaperID{4734} %
\def\confName{CVPR}
\def\confYear{2023}
\definecolor{armygreen}{rgb}{0.0, 0.5, 0.0}
\definecolor{sdvcolor}{rgb}{0.196, 0.431, 0.627}

\def \short {} %
\ifx \short \undefined
   \newcommand{\cutsectionup}{}
   \newcommand{\cutsectiondown}{}

   \newcommand{\cutsubsectionup}{}
   \newcommand{\cutsubsectiondown}{}

   \newcommand{\cutparagraphup}{}
   \newcommand{\cutparagraphdown}{}

   \newcommand{\cutcaptionup}{}
   \newcommand{\cutcaptiondown}{}

   \newcommand{\cuthalftablecaptionup}{}
   \newcommand{\cuthalftablecaptiondown}{}
\else
   \newcommand{\cutsectionup}{\vspace*{-2pt}}
   \newcommand{\cutsectiondown}{\vspace*{-2pt}}

   \newcommand{\cutsubsectionup}{\vspace*{-2pt}}
   \newcommand{\cutsubsectiondown}{\vspace*{2pt}}

   \newcommand{\cutparagraphup}{\vspace*{-3pt}}
   \newcommand{\cutparagraphdown}{\vspace*{-2pt}}

   \newcommand{\cutcaptionup}{\vspace*{-10pt}}
   \newcommand{\cutcaptiondown}{\vspace*{-10pt}}

   \newcommand{\cuthalftablecaptionup}{\vspace*{-10pt}}
   \newcommand{\cuthalftablecaptiondown}{\vspace*{-8pt}}
\fi

\newcommand{\name}{UniSim\xspace}
\newcommand{\fg}{features grid\xspace}
\newcommand{\simreal}{Sim2Real\xspace}
\newcommand{\realsim}{Real2Sim\xspace}

\makeatletter
\g@addto@macro\@maketitle{
\vspace{-3em}
\begin{figure}[H]
\setlength{\linewidth}{\textwidth}
\setlength{\hsize}{\textwidth}
\centering
\includegraphics[trim={0cm, 0cm, 0cm, 0.0cm},clip,width=\linewidth]{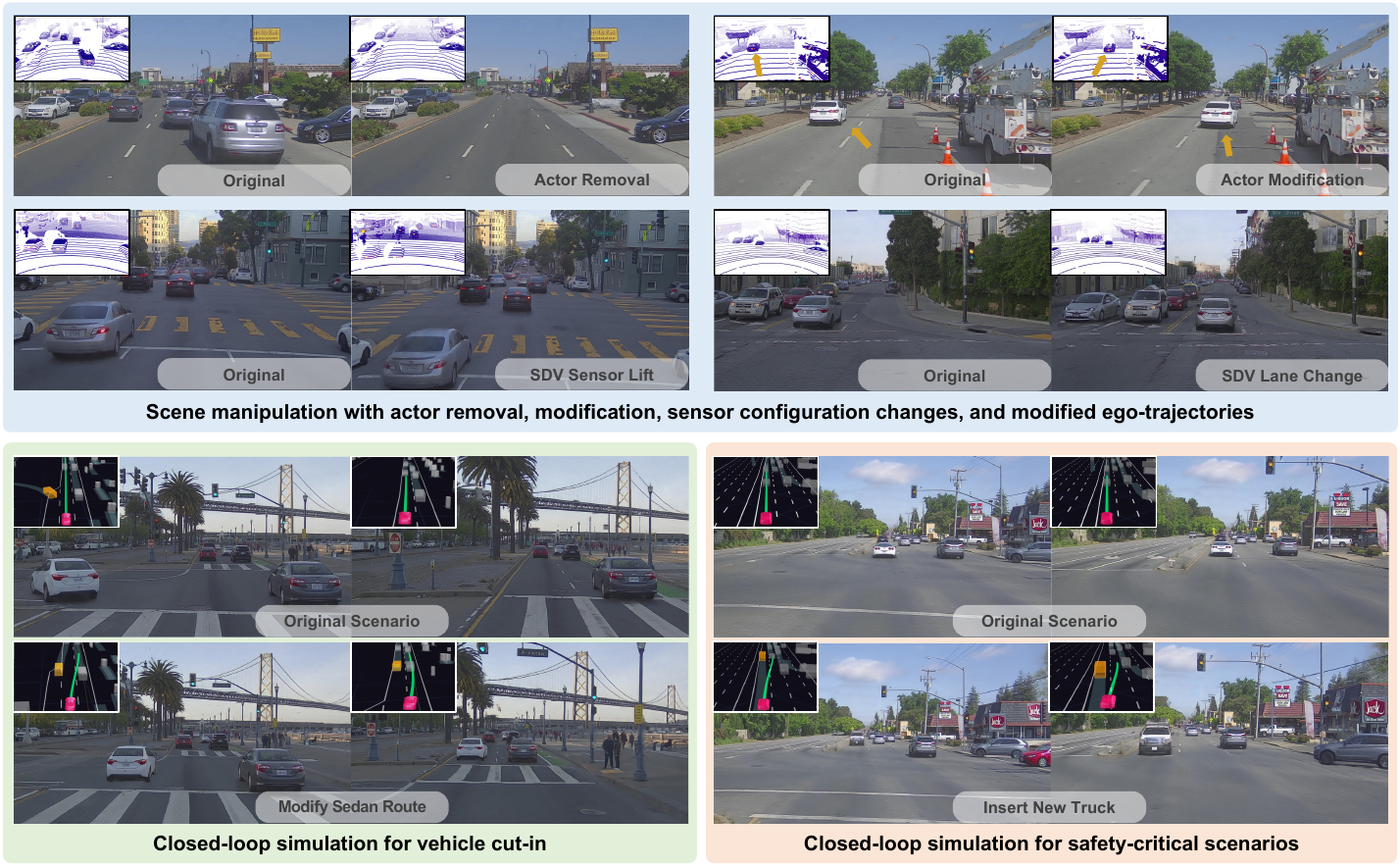}
\vspace{-6.5mm}
\caption{
	\textbf{Top:} \name takes recorded sensor data from a data collection platform and creates manipulable digital twins.
	\textbf{Bottom:} \name generates realistic, temporally consistent sensor simulations for new scenarios, enabling closed-loop autonomy evaluation.
	The autonomy system reactively interacts with the scenario, receives new sensor data, and changes lanes (\textcolor{armygreen}{see planned trajectory inset}).
	All images and LiDAR in figure simulated by \name.
	Please refer to our project page \url{https://waabi.ai/research/unisim/} for more results.
}
\label{fig:teaser}
\end{figure}
}
\makeatother

\begin{document}

\title{UniSim: A Neural Closed-Loop Sensor Simulator}

\author{
Ze Yang$^{1, 2}$\thanks{Indicates equal contribution.}
\quad Yun Chen$^{1, 2*}$
\quad Jingkang Wang$^{1, 2*}$
\quad Sivabalan Manivasagam$^{1, 2*}$ \\
\quad Wei-Chiu Ma$^{1, 3}$
\quad Anqi Joyce Yang$^{1, 2}$
\quad Raquel Urtasun$^{1, 2}$ \\
\normalsize{
$^{1}$Waabi
\quad $^{2}$University of Toronto
\quad $^{3}$Massachusetts Institute of Technology} \\
{\small\texttt{\{zyang,  ychen, jwang, siva, weichiu, jyang, urtasun\}@waabi.ai}}
}

\maketitle

\begin{abstract}
Rigorously testing autonomy systems is essential for making safe self-driving vehicles (SDV) a reality.
It requires one to generate safety critical scenarios beyond what can be collected safely in the world, as many scenarios happen rarely on public roads.
To accurately evaluate performance, we need to test the SDV on these scenarios in closed-loop, where the SDV and other actors interact with each other at each timestep.
Previously recorded driving logs provide a rich resource to build these new scenarios from, but for closed loop evaluation, we need to modify the sensor
data based on the new scene configuration and the SDV's decisions, as actors might be added or removed and the trajectories of existing actors and the SDV will differ from the original log.
In this paper, we present \name, a neural sensor simulator that takes a single recorded log captured by a sensor-equipped vehicle and converts it into a realistic closed-loop multi-sensor simulation.
\name builds neural feature grids to reconstruct both the static background and dynamic actors in the scene, and composites them together to simulate LiDAR and camera data at new viewpoints, with actors added or removed and at new placements.
To better handle extrapolated views, we incorporate learnable priors for dynamic objects, and leverage a convolutional network to complete unseen regions.
Our experiments show \name can simulate realistic sensor data with small domain gap on downstream tasks.
With \name, we demonstrate closed-loop evaluation of an autonomy system on safety-critical scenarios as if it were in the real world.
\end{abstract}

\cutcaptionup
\section{Introduction}
\cutsectiondown
\label{sec:intro}

While driving along a highway, a car from the left suddenly swerves into your lane.
You brake hard, avoiding an accident, but discomforting your passengers.
As you replay the encounter in your mind, you consider how the scenario would have gone if the other vehicle had accelerated more, if you had slowed down earlier, or if you had instead changed lanes for a more comfortable drive.
Having the ability to generate such ``what-if'' scenarios from a single recording would be a game changer for developing safe self-driving solutions.
Unfortunately, such a tool does not exist and the self-driving industry primarily test their systems on pre-recorded real-world sensor data (i.e., log-replay), or by driving ever more miles in the real-world.
In the former, the autonomous system cannot  execute actions and observe their effects as new sensor data different from the original recording is not generated, while the latter is neither safe, nor scalable or sustainable.
The status quo calls for novel closed-loop sensor simulation systems that are high fidelity and represent the diversity of the real world.

Here, we aim to build an editable digital twin of the real world (through the logs we captured), where existing actors in the scene can be modified or removed, new actors can be added, and new autonomy trajectories can be executed.
This enables the autonomy system to interact with the simulated world, where it receives new sensor observations based on its new location and the updated states of the dynamic actors, in a closed-loop fashion.
Such a simulator can accurately measure self-driving performance, as if it were actually in the real world, but without the safety hazards, and in a much less capital-intensive manner.
Compared to manually-created game-engine based virtual worlds~\cite{dosovitskiy2017carla,shah2018airsim}, it is a more scalable, cost-effective, realistic, and diverse way towards closed-loop evaluation.

Towards this goal, we present \name, a realistic closed-loop data-driven sensor simulation system for self-driving.
\name reconstructs and renders multi-sensor data for novel views and new scene configurations from a single recorded log.
This setting is very challenging as the observations are  sparse and often captured from constrained viewpoints (\eg, straight trajectories along the roads).
To better handle extrapolation from the observed views, we propose a series of enhancements over prior neural rendering approaches.
In particular, we leverage multi-resolution voxel-based neural fields to represent and compose the static scene and dynamic agents, and volume render feature maps.
To better handle novel views and incorporate scene context to reduce artifacts, a convolutional network (CNN) renders the feature map to form the final image.
For dynamic agents, we learn a neural shape prior that helps complete the objects to render unseen areas.
We use this sparse voxel-based representations to efficiently simulate both image and LiDAR observations under a unified framework.
This is very useful as SDVs often use several sensor modalities for robustness.

Our experiments show that \name realistically simulates camera and LiDAR observations at new views for large-scale dynamic driving scenes, achieving SoTA performance in photorealism.
Moreover, we find \name reduces the domain gap over existing camera simulation methods on the downstream autonomy tasks of detection, motion forecasting and motion planning.
We also apply \name to augment training data to improve perception models.
Importantly, we show, for the first time, closed-loop evaluation of an autonomy system on photorealistic safety-critical scenarios, allowing us to better measure SDV performance.
This further demonstrates \name's value in enabling safer and more efficient development of self-driving.

\section{Related Work}

\paragraph{Simulation Environments for Robotics:}
\cutparagraphdown
The robotics community has a long history of building simulators for safer and faster robot development~\cite{khosla1985parameter, wymann2000torcs, michel2004cyberbotics, diankov2008openrave, todorov2012mujoco, li2021igibson}.
Early works focused on modeling robot dynamics and physical forces for parameter identification and controller modelling~\cite{khosla1985parameter, neuman1985computational}.
Several works then developed accurate physics engines for improving robot design and motion planning~\cite{coumans2016pybullet, hugues2006simbad,diankov2008openrave, koenig2004design, brockman2016openai}, and for specific domains such as grasping~\cite{leon2010opengrasp}, soft robotics~\cite{hu2019chainqueen}, and SDVs~\cite{wymann2000torcs}.
But to enable end-to-end testing of full autonomy systems, we must also simulate realistic sensor observations of the 3D environment for the robot to perceive, interact with its surroundings, and plan accordingly~\cite{gibson2014ecological}.
Most prior sensor simulation systems use 3D-scanned or manually built synthetic environments for small indoor environments~\cite{koenig2004design, li2021igibson,savva2019habitat}, and perform rasterization or ray-tracing~\cite{shreiner2009opengl, parker2010optix} to simulate various sensor data~\cite{gschwandtner2011blensor, goodenough2012dirsig, karis2013real}.
For high-speed robots such as SDVs, simulators such as CARLA and AirSim~\cite{dosovitskiy2017carla, shah2018airsim} applied a similar approach.
But due to the costly manual effort in creating scenes, these simulators have difficulty scaling to all the areas we may want to test in, have limited asset diversity (e.g., roads, vehicles, vegetation) compared to the real world, and generate unrealistic sensor data that require substantial domain adaptation for autonomy~\cite{hoffman2018cycada, wu2019squeezesegv2}.

\cutparagraphup
\paragraph{Novel View Synthesis:}
\cutparagraphdown
Recent novel view synthesis (NVS) work has achieved success in automatically generating highly photorealistic sensor observations~\cite{riegler2020fvs, mildenhall2020nerf, ost2021neural, kundu2022panoptic, rematas2022urban, ost2022neural, aliev2020neural, lombardi2019neuralvolumes}.
Such methods aim to learn a scene representation from a set of densely collected observed images and render the scene from nearby unseen viewpoints.
Some works perform geometry reconstruction and then warp and aggregate pixel-features from the input images into new camera views, which are then processed by learning-based modules~\cite{riegler2020fvs, riegler2021svs, aliev2020neural, rakhimov2022npbg++}.
Others represent the scene implicitly as a neural radiance field (NeRF) and perform volume rendering with a neural network~\cite{mildenhall2020nerf,barron2021mip,verbin2022ref,yang2023neusim}.
These methods can represent complex geometry and appearance and have achieved photorealistic rendering, but focus on small static scenes.
Several representations~\cite{liu2020nsvf, chabra2020deep, martel2021acorn, neff2021donerf, rebain2021derf, reiser2021kilonerf, Tancik_2022_CVPR_blocknerf, mueller2022instant, zhang2020nerf++} partition the space and model the volume more efficiently to handle large-scale unbounded outdoor scenes.
However, these works focus primarily on the NVS task where a dense collection of images are available and most test viewpoints are close to the training views, and focus on the static scene without rendering dynamic objects such as moving vehicles.
In contrast, our work extends NVS techniques to build a sensor simulator from a single recorded log captured by a high-speed mobile platform.
We aim to render image and LiDAR observations of dynamic traffic scenarios from new viewpoints and modified scene configurations to enable closed-loop autonomy evaluation.

\cutparagraphup
\cutparagraphup
\paragraph{Data-driven Sensor Simulation for Self Driving:}
\cutparagraphdown
Several past works have leveraged computer vision techniques and real world data to build sensor simulators for self-driving.
Some works perform 3D reconstruction by aggregating LiDAR and building textured geometry primitives for physics-based rendering~\cite{tallavajhula2018off, manivasagam2020lidarsim, fang2021lidar, yang2020surfelgan}, but primarily simulate LiDAR or cannot model high-resolution images.
Another line of work perform object reconstruction and insertion into existing images~\cite{chen2021geosim, wang2022neural, yang2023neusim, wang2022cadsim} or point clouds~\cite{wang2021advsim, fang2020augmented, yang2021s3, yang2020recovering}, but these methods are unable to render sensor data from new views for closed-loop interaction.
DriveGAN~\cite{kim2021drivegan}  represents the scene as disentangled latent codes and generates video from control inputs with a neural network for differentiable closed-loop simulation, but is limited in its realism and is not temporally consistent.
AADS~\cite{li2019aads} and VISTA 2.0~\cite{amini2022vista, amini2020learning_vista, wang2022learning}, perform multi-sensor simulation via image-based warping or ray-casting on previously collected sensor data to render new views of the static scene, and then insert and blend CAD assets into the sensor data to create new scenarios.
These approaches, while promising, have visual artifacts for the inserted actors and rendered novel views, resulting in a large domain gap.
Neural Scene Graphs (NSG)~\cite{ost2021neural} and Panoptic Neural Fields (PNF)~\cite{kundu2022panoptic} represent the static scene and agents as multi-layer perceptrons (MLPs) and volume render photorealistic images of the scene.
However, the single MLP has difficulties modelling large scale scenes. 
These prior works also focus on scene editing and perception tasks where the SDV does not deviate significantly from the original recording.
Instead, we focus on multi-sensor simulation for closed loop evaluation of autonomy systems, and specifically design our system to better handle extrapolation.

\cutsectionup
\cutsectionup
\section{Neural Sensor Simulation}
\cutsectiondown

Given a log with camera images and LiDAR point clouds captured by a moving platform, as well as their relative poses in a reference frame, our goal is to construct an \emph{editable} and \emph{controllable} digital twin, from which we can generate realistic multi-modal sensor simulation and counterfactual scenarios of interest.
We build our model based on the intuition that the  3D world can be decomposed as a static background and a set of moving actors.
By effectively disentangling and modeling each component, we can manipulate the actors to generate new scenarios and simulate the sensor observations from new viewpoints.
Towards this goal, we propose \name, a neural rendering closed-loop simulator that jointly learns shape and appearance representations for both the static
scene and dynamic actors from the sensor data captured from a single pass of the environment.

\begin{figure}[t]
    \centering
     \includegraphics[width=0.9\linewidth]{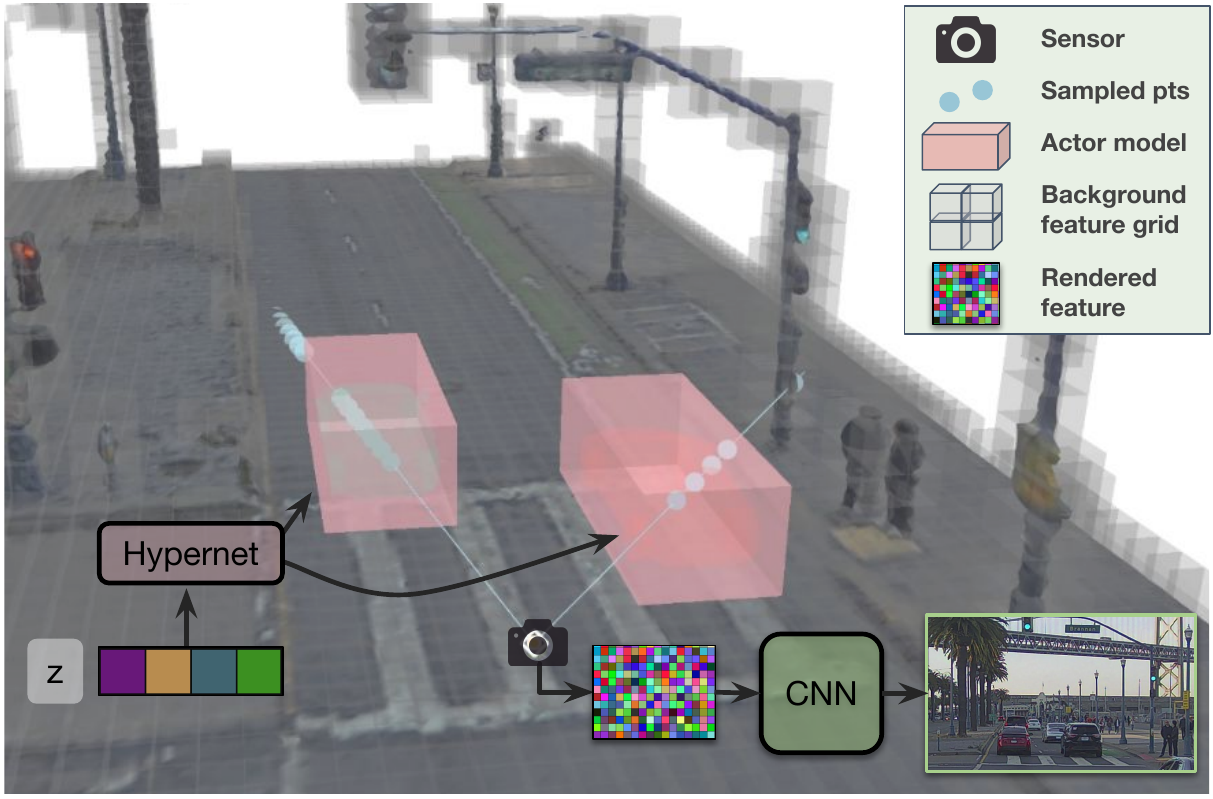}
     \cutcaptionup
     \vspace{5pt}
     \caption{\textbf{Overview of our approach:}
     We divide the 3D scene into a static background (grey) and a set of dynamic actors (red).
     We query the neural feature fields separately for static background and dynamic actor models, and perform volume rendering to generate neural feature descriptors.
     We model the static scene with a sparse feature-grid and use a hypernetwork to generate the representation of each actor from a learnable latent.
     We finally use a convolutional network to decode feature patches into an image.
     }
     \cutcaptiondown
     \label{fig:overview}
\end{figure}

We unfold this section by first reviewing the basic building blocks of our approach.
Next, we present our compositional scene representation, and detail how we design our background and dynamic actor models.
We then describe how to generate simulated sensor data with \name.
Finally, we discuss how to learn the model from real-world data.
Fig.~\ref{fig:overview} shows an overview of our approach.

\cutsubsectionup
\subsection{Preliminaries}
\cutsubsectiondown
\paragraph{Neural feature fields:}
\cutparagraphdown
A feature field refers to a continuous function $f$ that maps a 3D point $\mathbf x \in \mathbb R^3$ and a view direction $\mathbf d \in \mathbb R^2$ to an implicit geometry $s \in \mathbb R$ and a $N_f$-dimensional feature descriptor $\mathbf f \in \mathbb R^{N_f}$.
Since the function is often parameterized as a neural network $f_\theta: \mathbb R^3 \times \mathbb R^2 \rightarrow \mathbb R \times \mathbb R^{N_f}$, with $\theta$ the learnable weights, we call it neural feature field (NFF). 
NFFs can be seen as a superset of several existing works~\cite{mildenhall2020nerf, occupancynet}.
If we represent the implicit geometry as volume density $s \in \mathbb R^+$ and the feature descriptor as RGB radiance $\mathbf f \in \mathbb R^{3}$, NFFs become NeRFs~\cite{mildenhall2020nerf}.
If we enforce the implicit geometry to be the probability of occupancy, NFFs become occupancy functions~\cite{occupancynet}.
Importantly, NFFs naturally support composition~\cite{guo2020object, niemeyer2021giraffe, kundu2022panoptic}, enabling the combination of multiple relatively simple NFFs to form a complex neural field.

\cutparagraphup
\paragraph{Multi-resolution \fg:}
\cutparagraphdown
To improve the expressiveness and speed of NFFs, past works~\cite{takikawa2021nglod, mueller2022instant, chibane2020implicit, Yu2022MonoSDF} further combined learnable multi-resolution \fg $\{\mathcal G^l\}_{l=1}^L$ with a neural network $f$. 
Specifically, given a query point $\mathbf x \in \mathbb R^3$, the 3D feature grid at each level is first trilinearly interpolated.   
The interpolated features are then concatenated with the view direction $\mathbf d \in \mathbb R^2$, and the resulting features are processed with an MLP head to obtain the geometry $s$ and feature descriptor $\mathbf f$:
\begin{align}
    s, \mathbf f = f\left(\{\texttt{interp}(\mathbf x, \mathcal G^l)\}_{l=1}^L, \mathbf d \right).
    \label{eqn:feature_fileds}
\end{align}
These multi-scale features encode both global context and fine-grained details, providing richer information comparing to the original input $\mathbf x$. This also enables using a smaller $f$, which significantly reduces the inference time~\cite{sun2021direct,takikawa2021nglod}.
In practice, we optimize the \fg using a fixed number of features $\mathcal F$, and map the \fg $\{\mathcal G^l\}_{l=1}^L$ to $\mathcal F$ with a grid index hash function~\cite{mueller2022instant}.
Hereafter, we will use $\mathcal F$ and $\{\mathcal G^l\}_{l=1}^L$ interchangeably.

\cutsubsectionup
\subsection{Compositional Neural Scene Representation}
\cutsubsectiondown
We aim to build a compositional scene representation that best models the 3D world including the dynamic actors and static scene.
Given a recorded log captured by a data collection platform, we first define a 3D space volume over the traversed region. 
The volume consists of a static background $\mathcal B$ and a set of dynamic actors $\{\mathcal A_i\}_{i=1}^N$.
Each dynamic actor is parameterized as a bounding box of dimensions $\mathbf s_{\mathcal A_i} \in \mathbb R^3$, and its trajectory is defined by a sequence of SE(3) poses $\{\mathbf{T}_{\mathcal A_i}^{t}\}_{t=1}^{T}$.
We then model the static background and dynamic actors with separate multi-resolution \fg and NFFs.
Let the static background be expressed in the world frame.
We represent each actor in its object-centroid coordinate system (defined at the centroid of its bounding box), and transform their \fg to world coordinates to compose with the background.
This allows us to disentangle the 3D motion of each actor, and focus on representing shape and appearance.
To learn high-quality geometry~\cite{wang2021neus,yariv2020multiview}, we adopt the signed distance function (SDF) as our implicit geometry representation $s$.
We now describe each {component} in more detail.

\cutparagraphup
\paragraph{Sparse background scene model:}
\cutparagraphdown
We model the whole static scene with a multi-resolution \fg $\mathcal F_\text{bg}$ and an MLP head $f_\text{bg}$.
Since a self-driving log often spans hundreds to thousands of meters, it is computationally and memory expensive to maintain a dense, high-resolution voxel grid. 
We thus utilize geometry priors from LiDAR observations to identify near-surface voxels and optimize only their features.
Specifically, we first aggregate the static LiDAR point cloud from each frame to construct a dense 3D scene point cloud.
We then voxelize the scene point cloud and obtain a scene occupancy grid $\mathbf V_\text{occ}$. Finally, we apply morphological dilation to the occupancy grid and coarsely split the 3D space into free \vs near-surface space.
As the static background is often dominated by free space, this can significantly sparsify the \fg and reduce the computation cost.
The geometric prior also allows us to better model the 3D structure of the scene, which is critical when simulating novel viewpoints with large extrapolation.
To model distant regions, such as sky, we follow~\cite{zhang2020nerf++, barron2022mip} to extend our background scene model to unbounded scenes.

\cutparagraphup
\paragraph{Generalized actor model:}
\cutparagraphdown
One straightforward way to model the actors is to parameterize each actor $\mathcal A_i$ with a \fg $\mathcal F_{\mathcal A_i}$ and adopt a shared MLP head $f_{\mathcal A}$ for all actors. In this design, the individual \fg encodes instance-specific geometry and appearance, while the shared network maps them to the same feature space for downstream applications.
Unfortunately, such a design requires large memory for dense traffic scenes and, in practice, often leads to overfitting --- the \fg does not generalize well to unseen viewpoints.
To overcome such limitations, we propose to learn a hypernetwork~\cite{ha2016hypernetworks} over the parameters of all grids of features.
The intuition is that different actors are observed from different viewpoints, and thus their grids of features are informative in different regions.
By learning a prior over them, we can capture the correlations between the features and infer the invisible parts from the visible ones.
Specifically, we model each actor $\mathcal A_i$ with a low-dimensional latent code $\mathbf z_{\mathcal A_i}$ and learn a hypernetwork $f_{\mathbf z}$ to regress the \fg $\mathcal F_{\mathcal A_i}$:
\begin{align}
    \mathcal F_{\mathcal A_i} = f_{\mathbf z} (\mathbf z_{\mathcal A_i}).
    \label{eqn:hyper_network}
\end{align}
Similar to the background, we adopt a shared MLP head $f_{\mathcal A}$ to predict the geometry and feature descriptor at each sampled 3D point via Eq.~\ref{eqn:feature_fileds}.
We jointly optimize the actor latent codes $\{\mathbf z_{\mathcal A_i}\}$ during training.

\cutparagraphup
\paragraph{Composing neural feature fields:}
\cutparagraphdown
Inspired by works that {composite} solid objects \cite{guo2020object, ost2021neural} into a scene, we first transform object-centric neural fields of the foreground actors to world coordinates with the desired poses (\eg, using $\mathbf{T}_{\mathcal A_i}^{t}$ for reconstruction).
As the static background is a sparse \fg, we then simply replace the free space with the actor feature fields.
Through this simple operation, we can insert, remove, and manipulate the actors within the scene.

\cutsubsectionup
\subsection{Multi-modal Sensor Simulation}
\cutsubsectiondown
Now that we have a composed scene representation of the static and dynamic world, the next step is to render it into the data modality of interest.
In this work, we  focus on camera images and LiDAR point clouds, as they are the two main sensory modalities employed by modern SDVs.

\cutparagraphup
\paragraph{Camera simulation:}
\cutparagraphdown
Following recent success in view synthesis and generation~\cite{niemeyer2021giraffe, eg3d}, we adopt a hybrid volume and neural rendering framework for efficient photorealistic image simulation. 
Given a ray $\mathbf r(t) = \mathbf o + t\mathbf d$ shooting from the camera center $\mathbf o$ through the pixel center in direction $\mathbf d$, we first sample a set of 3D points along the ray and extract their features and geometry (Eq.~\ref{eqn:feature_fileds}). 
We then aggregate the samples and obtain a pixel-wise feature descriptor via volume rendering:
\begin{align}
    \mathbf f(\mathbf r) = \sum_{i=1}^{N_{\mathbf r}} w_i \mathbf f_i, \quad w_i = \alpha_i \prod_{j=1}^{i-1} (1-\alpha_j).
    \label{eqn:render_neural_feature}
\end{align}
Here, $\alpha_i \in [0, 1]$ represents opacity, which we can derive from the SDF $s_i$ using an approximate step function $\alpha = 1/(1 + \exp(\beta \cdot s))$, and  $\beta$ is the hyper-parameter controlling the slope.
We volume render all camera rays and generate a 2D feature map $\mathbf F \in \mathbb R^{H_f \times W_f \times N_f}$.
We then leverage a 2D CNN $g_\text{rgb}$ to \emph{render} the feature map to an RGB image $\mathbf I_\text{rgb}$:
\begin{align}
	 g_\text{rgb}: \mathbf F \in \mathbb R^{H_f \times W_f \times N_f} \rightarrow \mathbf I_\text{rgb} \in \mathbb R^{H \times W \times 3}.
	 \label{eqn:render_rgb}
\end{align}
In practice, we adopt a smaller spatial resolution for the feature map $H_f \times W_f$ than that of the rendered image $H \times W$, and rely on the CNN $g_\text{rgb}$ for upsampling. 
This allows us to significantly reduce the amount of ray queries.

\cutparagraphup
\paragraph{LiDAR simulation:}
\cutparagraphdown
LiDAR point clouds encode 3D (depth) and intensity (reflectivity) information, both of which can be simulated in a similar fashion to Eq.~\ref{eqn:render_neural_feature}.
We assume the LiDAR to be a time-of-flight pulse-based sensor, and model the pulses transmitted by the oriented LiDAR laser beams as a set of rays.
We slightly abuse the notation and let $\mathbf r(t) = \mathbf o + t\mathbf d$ be a ray casted from the LiDAR sensor we want to simulate. 
Denote $\mathbf o$ as the center of the LiDAR and $\mathbf d$ as the normalized vector of the corresponding beam.
We then simulate the depth measurement by computing the expected depth of the sampled 3D points:
\begin{align}
    D(\mathbf r) &= \sum_{i=1}^{N_{\mathbf r}} w_i t_i.
    \label{eqn:render_depth}
\end{align}
As for LiDAR intensity, we volume render the ray feature (using Eq.~\ref{eqn:render_neural_feature}) and pass it through an MLP intensity decoder $g_\text{int}$ to predict its intensity $l^\text{int}(\mathbf r) = g_\text{int}(\mathbf f(\mathbf r))$.

\cutsubsectionup
\subsection{Learning}
\cutsubsectiondown
\label{sec:learning}
We jointly optimize all grids of features $\mathcal F_{*}$ (including latent codes $\{\mathbf z_{\mathcal{A}_i}\}$, the  hypernetwork $f_{\mathbf z}$, the MLP heads ($f_\text{bg}, f_\mathcal{A}$) and the decoders ($g_\text{rgb}, g_\text{int}$) by minimizing the difference between the sensor observations and our rendered outputs.
We also regularize the underlying geometry such that it satisfies real-world constraints. 
Our full objective is:
\begin{align*}
   \mathcal{L} = \mathcal{L}_\text{rgb} + \lambda_\text{lidar}\mathcal{L}_\text{lidar} + \lambda_\text{reg}\mathcal{L}_\text{reg} +  \lambda_\text{adv}\mathcal{L}_\text{adv}.
\end{align*}
In the following, we discuss in more detail each term.

\vspace{-3pt}
\cutparagraphup
\paragraph{Image simulation $\mathcal L_\text{rgb}$:}
\cutparagraphdown
This objective consists of a $\ell_2$ photometric loss and a perceptual loss~\cite{zhang2018unreasonable, wang2018pix2pixHD}, both measured between the observed images and our simulated results. 
We compute the loss in a patch-wise fashion:
\begin{equation}
\resizebox{.43\textwidth}{!}{$
\begin{split}
	\mathcal L_\text{rgb} = \frac{1}{N_{\text{rgb}}} \sum\limits_{i=1}^{N_{\text{rgb}}} \left( \left\lVert \mathbf I^{\text{rgb}}_i - \mathbf{ \hat {I}}^{\text{rgb}}_i \right\rVert_2 +
    \lambda \sum_{j=1}^M \left\lVert V^j(\mathbf I^{\text{rgb}}_i) - V^j(\mathbf{ \hat {I}}^{\text{rgb}}_i) \right\rVert_1 \right),
\end{split}$}
\end{equation}
where $\mathbf I^{\text{rgb}}_i = f_\text{rgb}(\mathbf F_i)$ is the rendered image patch (Eq.~\ref{eqn:render_rgb}) and $\mathbf{ \hat {I}}^{\text{rgb}}_i$ is the corresponding observed image patch.
$V^j$ denotes the $j$-th layer of a pre-trained VGG network~\cite{simonyan2014very}.

\vspace{-3pt}
\cutparagraphup
\paragraph{LiDAR simulation $\mathcal L_\text{lidar}$:}
\cutparagraphdown
This objective measures the $\ell_2$ error between the observed LiDAR point clouds and the simulated ones. Specifically, we compute the depth  and  intensity differences:
\begin{equation}
\resizebox{.43\textwidth}{!}{$
\begin{split}
    \mathcal L_\text{lidar} = \frac{1}{N} \sum\limits_{i=1}^N \left(\left\lVert D(\mathbf r_i) - D^{\text{obs}}_i \right\rVert_2  +  \left\lVert l^{\text{int}}(\mathbf r_i) - \hat{l}^{\text{int}}_i \right\rVert_2 \right).
\end{split}$}
\end{equation}
Since LiDAR observations are noisy, we filter outliers and encourage the model to focus on credible supervision.
In practice, we optimize 95\% of the rays within each batch that have smallest depth error.

\vspace{-3pt}
\cutparagraphup
\paragraph{Regularization $\mathcal L_\text{reg}$:}
\cutparagraphdown
We further apply two additional constraints on the learned representations. 
First, we encourage the learned sample weight distribution $w$ (Eq.~\ref{eqn:render_neural_feature}) to concentrate around the surface.
Second, we encourage the underlying SDF $s$ to satisfy the Eikonal equation, which helps the network optimization find a smooth zero level set~\cite{gropp2020implicit}:
\begin{equation}
\label{eqn:total_loss}
\resizebox{.43\textwidth}{!}{$
\begin{split}
    \mathcal L_\text{reg} = \frac{1}{N} \sum\limits_{i=1}^N  \biggl( \sum\limits_{\tau_{i,j} > \epsilon} \left\lVert w_{ij} \right\rVert_2  + \sum\limits_{\tau_{i,j} < \epsilon} \left( \left\lVert \nabla s(\mathbf x_{ij}) \right\rVert_2 - 1 \right)^2 \biggr),
\end{split}$}
\end{equation}
where $\tau_{i,j} = |t_{ij} - {D}^{\text{gt}}_i|$ is the distance between the sample $\mathbf x_{ij}$ and its corresponding LiDAR observation ${D}^{\text{gt}}_i$.

\begin{figure*}[ht]
    \centering
     \includegraphics[trim={0 0 0 0},clip, width=1.0\linewidth]{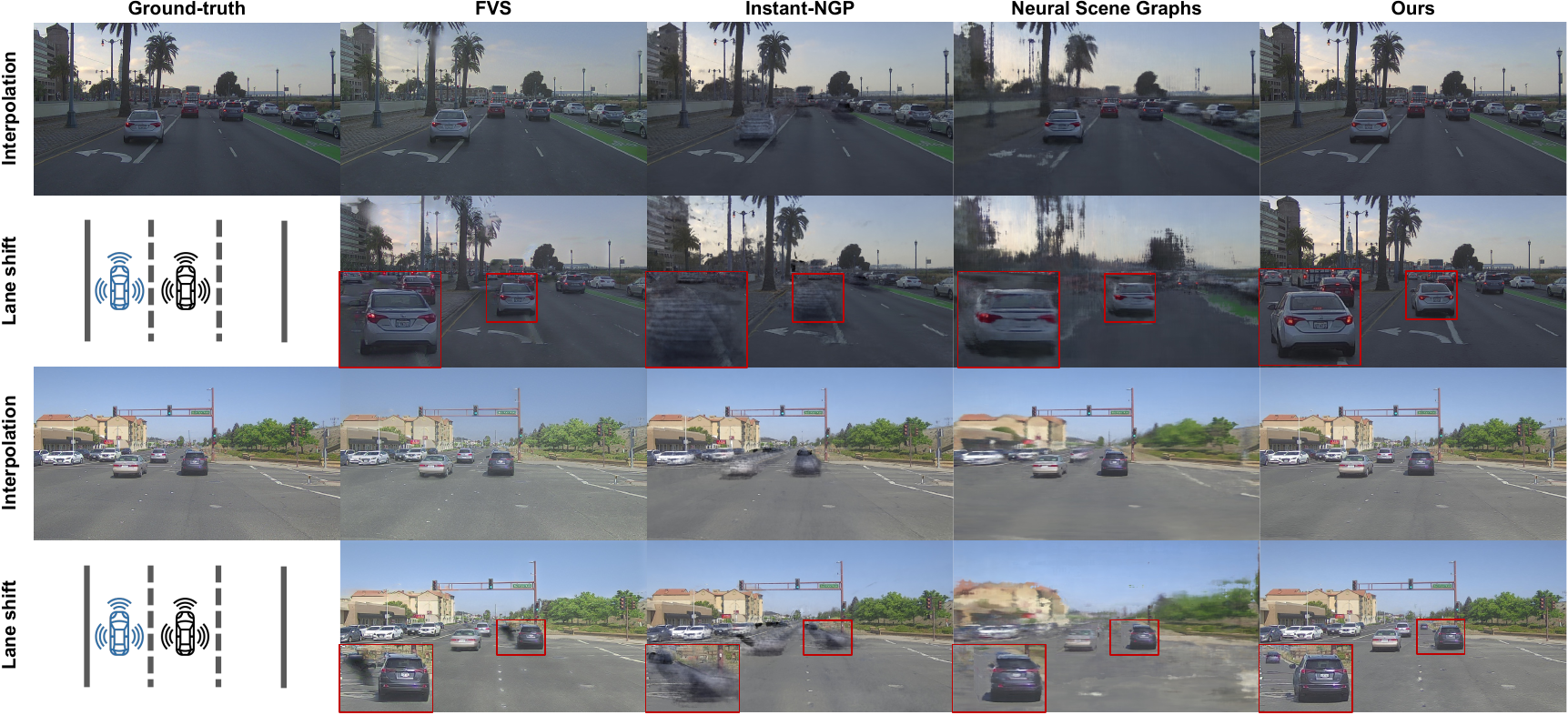}
     \cutcaptionup
     \cutcaptionup
     \caption{\textbf{Qualitative comparison.} 
     We show simulation results in both the interpolation (rows 1, 3) and \textcolor{sdvcolor}{lane-shift} test settings (rows 2, 4).
     }
     \cutcaptiondown
     \label{fig:panda_qual}
\end{figure*}

\begin{table}[]
    \centering
    \resizebox{0.5\textwidth}{!}{
    \begin{tabular}{lccccc}
    \toprule
    \multirow{2}{*}{Methods} & \multicolumn{3}{c}{Interpolation} &
    \multicolumn{2}{c}{Lane Shift} \\
    \cmidrule(r){2-4} \cmidrule(l){5-6}
    & {PSNR$\uparrow$ } & {SSIM$\uparrow$ } & {LPIPS$\downarrow$ } & {FID$\downarrow $ @ 2m} & {FID$\downarrow $ @ 3m} \\
    \midrule
    FVS~\cite{riegler2020fvs} & 21.09 & 0.700 & 0.299 & \multicolumn{1}{r}{112.6} & \multicolumn{1}{r}{135.8} \\
    NSG~\cite{ost2021neural} & 20.74 & 0.600 & 0.556 & \multicolumn{1}{r}{319.2} & \multicolumn{1}{r}{343.0} \\
    Instant-NGP~\cite{mueller2022instant} & 24.03 & 0.708 & 0.451 & \multicolumn{1}{r}{192.8} & \multicolumn{1}{r}{220.1} \\
    Ours & \textbf{25.63} & \textbf{0.745} & \textbf{0.288} & \multicolumn{1}{r}{\textbf{74.7}} & \multicolumn{1}{r}{\textbf{97.5}} \\
    \bottomrule
    \end{tabular}}
	\cuthalftablecaptionup
	\caption{\textbf{State-of-the-art image comparison}}
	\cuthalftablecaptiondown
	\label{tab:sota_panda}
\end{table}

\begin{table}[]
    \centering
    \resizebox{0.5\textwidth}{!}{
    \begin{tabular}{lccccc}
    \toprule
    \multirow{2}{*}{Methods} & \multicolumn{3}{c}{Interpolation} & \multicolumn{2}{c}{Lane Shift} \\
    \cmidrule(r){2-4} \cmidrule(l){5-6}
    & {PSNR$\uparrow$ } & {SSIM$\uparrow$ } & {LPIPS$\downarrow$ } & {FID$\downarrow $ @ 2m} & {FID$\downarrow $ @ 3m} \\
    \midrule
    NFF only & 24.93 & 0.717 & 0.393 & \multicolumn{1}{r}{153.7} & \multicolumn{1}{r}{173.5} \\
    + Actor model & 25.80 & 0.744 & 0.364 & \multicolumn{1}{r}{84.1} & \multicolumn{1}{r}{111.8} \\
    + CNN & \textbf{25.99} & \textbf{0.762} & 0.341 & \multicolumn{1}{r}{78.8} & \multicolumn{1}{r}{103.3} \\
    + VGG \& GAN loss & 25.63 & 0.745 & \textbf{0.288} & \multicolumn{1}{r}{\textbf{74.7}} & \multicolumn{1}{r}{\textbf{97.5}} \\
    \bottomrule
    \end{tabular}}
	\cuthalftablecaptionup
	\caption{\textbf{Ablation of \name enhancements}
    }
	\cuthalftablecaptiondown
	\label{tab:ablate}
\end{table}

\begin{table}[]
	\centering
	\resizebox{0.5\textwidth}{!}{
		\begin{tabular}{@{}lccc@{}}
			\toprule
			& Median $\ell_2$ Error (m)$\downarrow$ & Hit Rate$\uparrow$ & Intensity RMSE$\downarrow$ \\ 
			\midrule
			LiDARsim~\cite{manivasagam2020lidarsim} & 0.11 & 92.2\% & 0.091 \\
			Ours & \textbf{0.10} & \textbf{99.6\%} & \textbf{0.065} \\ 
			\bottomrule
		\end{tabular}
	}
	\cuthalftablecaptionup
	\caption{\textbf{State-of-the-art LiDAR comparison}}
	\cuthalftablecaptiondown
	\label{tab:sota_panda_lidar}
\end{table}

\cutparagraphup
\paragraph{Adversarial loss $\mathcal L_\text{adv}$:}
\cutparagraphdown
To improve photorealism at unobserved viewpoints, we train a discriminator CNN $\mathcal D_{\text{adv}}$ to differentiate between our simulated images at observed viewpoints and unobserved ones.
Specifically, we denote the set of rays to render an image patch as $\mathbf R = \{\mathbf r(\mathbf o, \mathbf d_j)\}_{j=1}^{P \times P}$, and randomly jitter the ray origin to create unobserved ray patches $\mathbf R' = \{\mathbf r(\mathbf o + \epsilon, \mathbf d_j)\}_{j=1}^{P \times P}$, where $\epsilon \in \mathcal N(0, \sigma)$.
The discriminator CNN $\mathcal D_{\text{adv}}$ minimizes:
\vspace{-0.05in}
\begin{align}
	- \frac{1}{N_\text{adv}} \sum_{i=1}^{N_\text{adv}} \log \mathcal D_{\text{adv}}(\mathbf I_i^{\text{rgb}, \mathbf R}) + \log (1- \mathcal D_{\text{adv}}(\mathbf I_i^{\text{rgb}, \mathbf R'})),
\end{align}
where $\mathbf I_i^{\text{rgb}, \mathbf R} = f_\text{rgb}(\mathbf F (\mathbf R_i))$ and $\mathbf I_i^{\text{rgb}, \mathbf R'} = f_\text{rgb}(\mathbf F (\mathbf R'_i))$ are the rendered image patches at observed and unobserved viewpoints, respectively.
We then define the adversarial loss $\mathcal L_\text{adv}$ to train the CNN RGB decoder $g_\text{rgb}$ and neural feature fields to improve photorealism at unobserved viewpoints as:
\vspace{-0.15in}
\begin{align}
	\mathcal L_\text{adv} = \frac{1}{N_\text{adv}} \sum_{i=1}^{N_\text{adv}} \log (1- \mathcal D_{\text{adv}}(\mathbf I_i^{\text{rgb}, \mathbf R'})).
\end{align}
\vspace{-0.25in}

\cutparagraphup
\paragraph{Implementation details:}
\cutparagraphdown
We identify actors along rendered rays using the AABB ray-box intersection~\cite{majercik2018ray}.
When sampling points along the ray, we adopt a larger step size for background regions and a smaller one for intersected actor models to ensure appropriate resolution.
We leverage the scene occupancy grid $\mathbf V_\text{occ}$ to skip point samples in free space. 
During learning, we also optimize the actor trajectories to account for noise in the initial input.
For vehicle actors, we also leverage the shape prior that they are symmetric along their length.

\cutsectionup
\section{Experiments}
\cutsectiondown

In this section we begin by introducing our experimental setting, and then compare our model against state-of-the-art methods to evaluate the sensor simulation realism and domain gap with real data, and also ablate our model components.
We then show that our method can generate diverse sensor simulations to improve vehicle detection.
Finally, we demonstrate \name for evaluating an autonomy system trained only on real data in closed-loop.

\vspace{2pt}
\cutsubsectionup
\subsection{Experimental Details}
\cutsubsectiondown
\vspace{2pt}

\cutparagraphup
\paragraph{Dataset:}
\cutparagraphdown
We evaluate our method on the publicly available PandaSet~\cite{xiao2021pandaset} which contains 103 driving scenes captured in urban areas in San Francisco.
Each scene is composed of 8 seconds (80 frames, sampled at 10hz) of images captured from a front-facing wide angle camera (1920$\times$1080) and point clouds from $360^\circ$ spinning LiDAR.

\cutparagraphup
\paragraph{Baselines:}
\cutparagraphdown
We compare our model against several SoTA methods.
\textbf{FVS}~\cite{riegler2020fvs} is an NVS method that uses reconstructed geometry (aggregated LiDAR in our implementation) as a ``proxy'' to re-project pixels from the input images into new camera views, where they are blended by a neural network.
We enhance FVS to model dynamic actors.
\textbf{Instant-NGP}~\cite{mueller2022instant} is a NeRF-based method that adopts multi-resolution hashing encoding for compact scene representation and efficient rendering.
We enhance it by adding LiDAR depth supervision for better geometry and extrapolation.
\textbf{NSG}~\cite{ost2021neural} is a camera simulation method that models the scene with separate NeRF representations for the static background and each dynamic actor.
See supp. for details.

\cutsubsectionup
\subsection{\name Controllability}
\cutsubsectiondown
We first highlight in Fig.~\ref{fig:teaser} the power of \name to perform all the capabilities for closed-loop sensor simulation.
We can not only render the original scene, but because of our decomposed actor and background representations, we can also remove all the actors, and change their positions.
With enhanced extrapolation capabilities, we can change the SDV's location or test new sensor configurations.
See supp. for more results, including highway scenes.

\cutsubsectionup
\subsection{Realism Evaluation}
\cutsubsectiondown
\vspace{-5pt}
Sensor simulation should not only reconstruct nearby views, but also generate realistic data at significantly different viewpoints.
Here we evaluate both settings.
Similar to other NVS benchmarks~\cite{liao2022kitti}, we subsample the sensor data by two, training on every other frame and testing on the remaining
frames, dubbed \emph{``interpolation''} test.
We report PSNR, SSIM~\cite{wang2004image}, and LPIPS~\cite{zhang2018unreasonable}.
We also evaluate extrapolation by simulating a new trajectory shifted laterally to the left or right by 2 or 3 meters, dubbed \emph{``lane shift''} test.
Since ground-truth is unavailable, we report FID~\cite{heusel2017gans}.

\begin{figure}[t]
    \centering
     \includegraphics[width=1.0\linewidth]{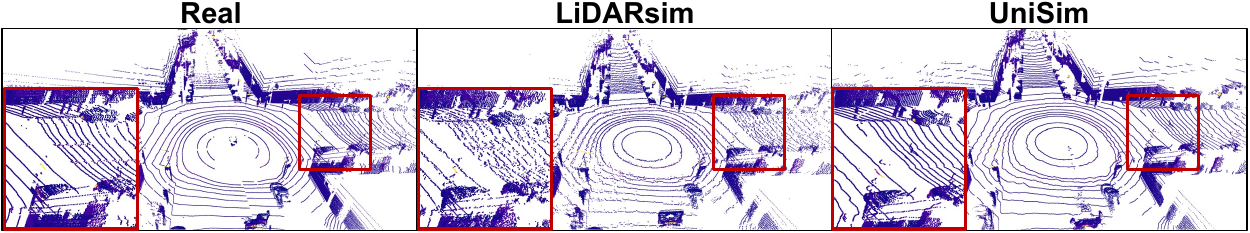}
     \cutcaptionup
     \vspace{-10pt}
	 \caption{\textbf{Comparison of LiDAR simulation.}
		\name produces higher-fidelity LiDAR simulation with less noise and more continuous beam rings that are closer to real LiDAR compared to~\cite{manivasagam2020lidarsim}.
	 }
     \cutcaptiondown
     \cutcaptiondown
     \label{fig:lidarsim_qual}
\end{figure}

\cutparagraphup
\paragraph{Camera Simulation:}
\cutparagraphdown
We report image-similarity metrics against SoTA in Table~\ref{tab:sota_panda}.
Due to computational costs of the baseline NSG, we select 10 scenes for evaluation.
Our method outperforms the baselines in all metrics, and the gap is more significant in extrapolation settings.
FVS performs well on LPIPS and InstantNGP on PSNR in the interpolation setting, but both have difficulty when rendering at extrapolated views.
Fig.~\ref{fig:panda_qual} shows qualitative results.
NSG produces decent results for dynamic actors but fails on large static scenes, due to its sparse multi-plane representation.
Note \name is more realistic than the baselines.

\cutparagraphup
\paragraph{Ablation:}
\cutparagraphdown
We validate the effectiveness of several key components in Tab.~\ref{tab:ablate}.
Both the actor model and the CNN decoder improve the overall performance over the neural \fg base model.
The CNN is especially effective in the extrapolation setting, as it improves the overall image quality by spatial relation reasoning and increases model capacity.
Adding perceptual and adversarial losses results in a small performance drop for interpolation, but improves the lane shift results.
Please see supp. for more visual results.

\vspace{-5pt}
\cutparagraphup
\paragraph{LiDAR Simulation:}
\cutparagraphdown
We also evaluate the fidelity of our LiDAR simulation and compare with SoTA approach LiDARsim~\cite{manivasagam2020lidarsim}.
For LiDARsim, we reconstruct surfel assets using all training frames, place actors in their original scenario in test frames, and perform ray-casting.
Both methods use the real LiDAR rays to generate a simulated point cloud.
We evaluate the fraction of real LiDAR points that have a corresponding simulated point (\ie, Hit rate), the median per-ray $\ell_2$ error and the average intensity simulation errors.
As shown in Tab.~\ref{tab:sota_panda_lidar}, \name outperforms LiDARsim in all metrics suggesting it is more accurate and has better coverage.
Fig.~\ref{fig:lidarsim_qual} shows a visual comparison.
Please see supp. for additional autonomy results and qualitative examples.

\begin{figure}[t]
    \centering
     \includegraphics[width=1.0\linewidth]{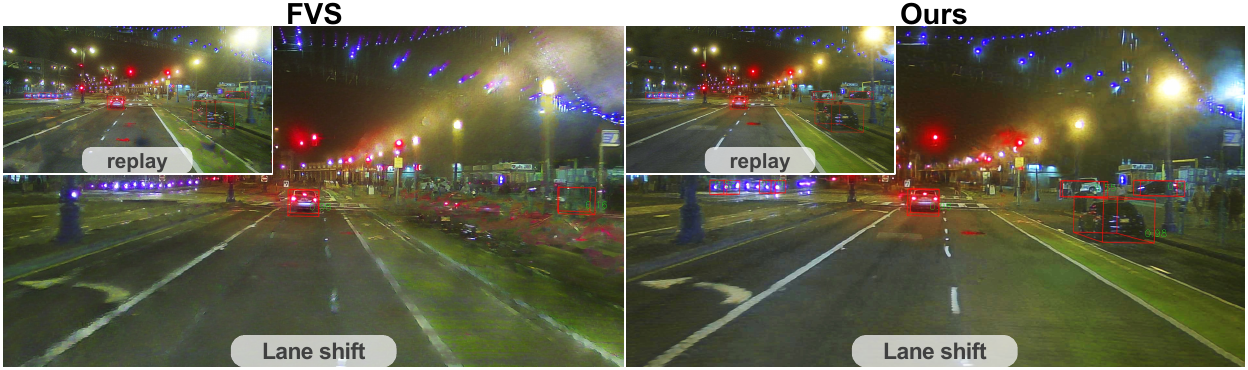}
     \cutcaptionup
     \vspace{-10pt}
     \caption{\textbf{Real2Sim Qualitative} on replay and lane shift settings.
     }
     \cutcaptiondown
     \label{fig:real2sim}
\end{figure}

\begin{table}[t]
	\centering
	\resizebox{0.45\textwidth}{!}{
		\begin{tabular}{@{}llllll@{}}
			\toprule
             & \multicolumn{2}{c}{Log Replay} & \multicolumn{2}{c}{Lane Shift} \\
			\cmidrule(r){2-3} \cmidrule(l){4-5}
			Method & \multicolumn{1}{c}{\realsim} & \multicolumn{1}{c}{\simreal} & \multicolumn{1}{c}{\realsim} & \multicolumn{1}{c}{\simreal} \\
			\midrule
			FVS~\cite{riegler2020fvs} & $36.9$ & $38.7$ & $30.3$ & $32.2$ \\
			Instant-NGP~\cite{mueller2022instant} & $22.6$  & $34.0$ & $18.1$ & $26.5$  \\
			Ours & $\mathbf{40.2}$ & $\mathbf{39.9}$ & $\mathbf{37.0}$ & $\mathbf{37.1}$ \\
			\bottomrule
		\end{tabular}
	}
	\cuthalftablecaptionup
	\caption{\textbf{Detection domain gap, mAP.} Real2Real = $40.9$.}
	\cuthalftablecaptiondown
	\label{tab:downstream}
\end{table}

\begin{table}[]
    \centering
    \resizebox{0.45\textwidth}{!}{
        \begin{tabular}{@{}lccc@{}}
            \toprule
             & Instant-NGP~\cite{mueller2022instant} & FVS~\cite{riegler2020fvs} &  Ours \\
            \midrule
            Sim & $32.4$ & $39.2$ & $\mathbf{41.4}$ \\
            Real + Sim & $40.1$  & ${41.1}$ & $\mathbf{42.9}$ \\
            \bottomrule
        \end{tabular}
    }
    	\cuthalftablecaptionup
    	\vspace{0.04in}
        \caption{\textbf{Augmenting with simulation, mAP.} Real2Real = $40.9$.}
        \cuthalftablecaptiondown
        \label{tab:aug}
\end{table}

\begin{table}[]
	\centering
	\resizebox{0.45\textwidth}{!}{
		\begin{tabular}{@{}lccc@{}}
			\toprule
			 & Det. Agg. $\uparrow$ & Pred. ADE $\downarrow$ & Plan Cons. $\downarrow$ \\
			\midrule
			FVS~\cite{riegler2020fvs} & $0.80$ & $2.35$ & \multicolumn{1}{r}{$6.15$} \\
			Instant-NGP~\cite{mueller2022instant} & $0.42$ & $3.24 $ & \multicolumn{1}{r}{$13.44 $} \\
			Ours & $\mathbf{0.82}$ & $\mathbf{1.68}$ & \multicolumn{1}{r}{$\mathbf{6.09}$} \\
			\midrule
		\end{tabular}
	}
	\cuthalftablecaptionup
	\caption{\textbf{Open-Loop \realsim Autonomy Evaluation}}
	\cuthalftablecaptiondown
	\vspace{-12pt}
	\label{tab:downstream_autonomy_interp}
\end{table}

\begin{figure*}[ht]
    \centering
    \includegraphics[trim={0 0 0 0},clip, width=0.99\linewidth]{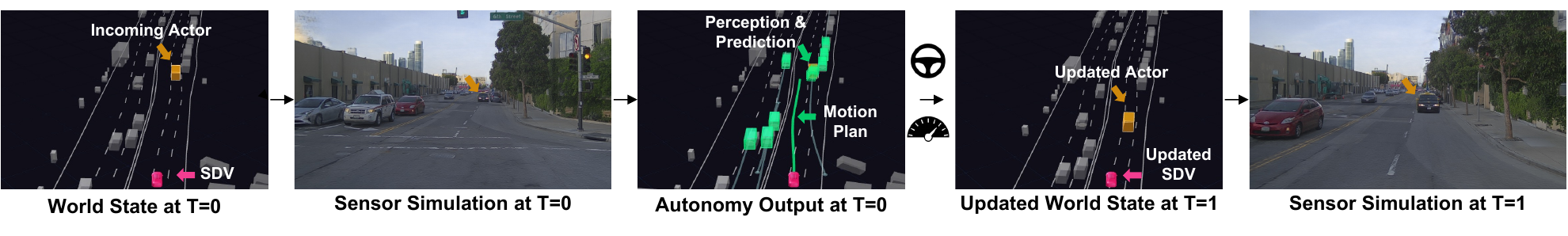}
    \cutcaptionup
    \caption{\textbf{Closed-loop Evaluation}.
    From left to right: With \name, we can create a safety-critical scenario (e.g., \textcolor{orange}{incoming actor}), simulate the sensor data, \textcolor{armygreen}{run autonomy} on it, update the SDV's viewpoint and other actor locations, and simulate the new sensor data.
    }
    \cutcaptiondown
    \vspace{-5pt}
    \label{fig:close_loop}
\end{figure*}

\cutsubsectionup
\subsection{Perception Evaluation and Training}
\cutsubsectiondown
In addition to image-similarity, sensor simulation should be realistic with respect to how autonomy perceives it.
To verify if UniSim reduces the domain gap for perception tasks, we leveraged the SoTA camera-based birds-eye-view (BEV) detection model BEVFormer~\cite{li2022bevformer}.
We consider two setups
(a) \textbf{\realsim}: evaluating the perception model trained on real data on simulated data;
(b) \textbf{\simreal}: training perception models with simulated data and testing on real data.
Specifically, we evaluate the real model on $24$ simulated validation logs for Real2Sim and train perception models with $79$ simulated training logs for Sim2Real.

We consider both \textit{replay} and \textit{lane shift} test settings.
In \textit{replay}, we replay all actors and SDV with their original trajectories.
In \textit{lane shift}, we shift the SDV trajectory laterally by $2$ meters and simulate images at extrapolated views.
We report detection mean average precision (mAP).

\cutparagraphup
\cutparagraphup
\paragraph{Domain Gap in Simulation:}
\cutparagraphdown
As shown in Table~\ref{tab:downstream}, our approach achieves the smallest domain gap in both \realsim and \simreal setups, on both \textit{replay} and \textit{lane shift} settings, while other existing approaches result in larger domain gaps, hindering their applicability to evaluate and train autonomy.
This is especially evident in the more challenging \textit{lane shift} setting, where there is a larger performance gap between \name and the baselines.
Fig.~\ref{fig:real2sim} shows the \realsim detection performance for both replay and lane shift settings compared to FVS~\cite{riegler2020fvs}.

\vspace{-1pt}
\cutparagraphup
\paragraph{Data Augmentation with Simulation Data:}
\cutparagraphdown
We now study if our simulated data boosts performance when used for training.
Specifically, we use all PandaSet training logs to generate simulation variations (replay, lane shift 0.5 and 2 meters) to train the detectors.
As shown in Table~\ref{tab:aug}, using \name data only to train the perception model is even better than training with all real data.
Note we only increase the rendered viewpoints and do not alter the content.
We then combine the real data with the simulation data and retrain the detector.
Table~\ref{tab:aug} shows \name augmentation yields a significant performance gain.
In contrast, baseline data augmentation brings marginal gain or harms performance.

\cutsubsectionup
\subsection{Full Autonomy Evaluation with \name}
\cutsubsectiondown

\cutparagraphup
\paragraph{Domain gap evaluation:}
\cutparagraphdown
Sensor simulation not only affects perception tasks, but also downstream tasks such as motion forecasting and planning.
We report domain gap metrics by evaluating an autonomy system trained on real data on simulated images of the original scenario.
The autonomy system under evaluation is a module-based system, with BEVFormer~\cite{li2022bevformer} taking front-view camera images as input and producing BEV detections that are matched over time to produce tracks via greedy association as the perception module.
These are then fed to a motion forecasting model~\cite{cui2018multimodal} that takes in BEV tracks and a map raster and outputs bounding boxes and 6 second trajectory forecasts.
Finally a SoTA sampling-based motion planner~\cite{plt} takes the prediction output and map to plan a maneuver.
We report open-loop autonomy metrics (detection agreement @ IoU 0.3, prediction average displacement error (ADE), and motion plan consistency at 5 seconds) in Table~\ref{tab:downstream_autonomy_interp}.
Compared to other methods, our approach has the smallest domain gap.
Please see supp. for details.

\vspace{-0.1in}
\cutparagraphup
\paragraph{Closed-loop Simulation:}
\cutparagraphdown
With \name, we can create new scenarios, simulate the sensor data, run the autonomy system, update the state of the actors in a reactive manner and the SDV's location, and execute the next time step (see Fig.~\ref{fig:close_loop}).
This gives us a more accurate measure of the SDV's performance to how it would behave in the real world for the same scenario.
Fig. \ref{fig:teaser} shows additional simulations of the autonomy on safety critical scenarios such as an actor cutting into our lane or an oncoming actor in our lane.
The SDV then lane changes, and with \name we can simulate the sensor data realistically throughout the scenario.
Please see supp. video for complete visuals.

\cutsectionup
\section{Conclusion}
\cutsectiondown

In this paper, we leveraged real world scenarios collected by a mobile platform to build a high-fidelity virtual world for autonomy testing.
Towards this goal, we presented \name, a neural sensor simulator that takes in a sequence of LiDAR and camera data and can decompose and reconstruct the dynamic actors and static background in the scene, allowing us to create new scenarios and render sensor observations of those new scenarios from new viewpoints.
\name improves over SoTA and generates realistic sensor data with much lower domain gap.
Furthermore,  we demonstrated that we can use it to evaluate an autonomy system in closed loop on novel safety-critical scenarios.
Such a simulator can accurately measure self-driving performance, as if it were actually in the real world, but without the safety hazards, and in a much less capital-intensive manner.
{We hope \name will enable developing safer autonomy systems more efficiently and safely.}
Future work involves explicitly modelling and manipulating scene lighting \cite{srinivasan2021nerv, boss2021neural, zhang2021nerfactor}, weather~\cite{li2022climatenerf}, and articulated actors~\cite{wang2022cadsim}.

\vspace{-20pt}
\paragraph{Acknowledgements:}
We thank  Ioan-Andrei Barsan for profound discussion and constructive feedback.
We thank the Waabi team for their valuable assistance and support.

{\small
\bibliographystyle{ieee_fullname}
\bibliography{egbib}
}

\clearpage

\onecolumn
{\hspace{-5mm} \LARGE{\textbf{Appendix}}\\}
\maketitle
\appendix
\renewcommand{\thesection}{A\arabic{section}}
In the supplementary material, we provide implementation details, experiment setting details, additional qualitative visualizations and metrics, and limitations of our method.
We first provide details about \name in Sec.~\ref{sec:additional_details}.
Then in Sec.~\ref{sec:baseline_details} we provide additional details on the baseline model implementations and how we adapt them to our sensor simulation setting.
In Sec.~\ref{sec:exp_details} we explain in more detail our experimental setting for the experiments in the main paper.
Next, we showcase additional visualizations and metrics for camera and LiDAR in Sec.~\ref{sec:additional_results}.
Finally, we analyze the limitations of our model in Sec.~\ref{sec:limitation}.
Please refer to our project page \url{https://waabi.ai/research/unisim/} for an overview of our methodology and video results on the various capabilities of \name and closed-loop autonomy simulations.

\section{\name Details}
\label{sec:additional_details}

\paragraph{Scene Representation Details:}

\begin{figure*}[ht]
    \centering
     \includegraphics[width=0.9\linewidth]{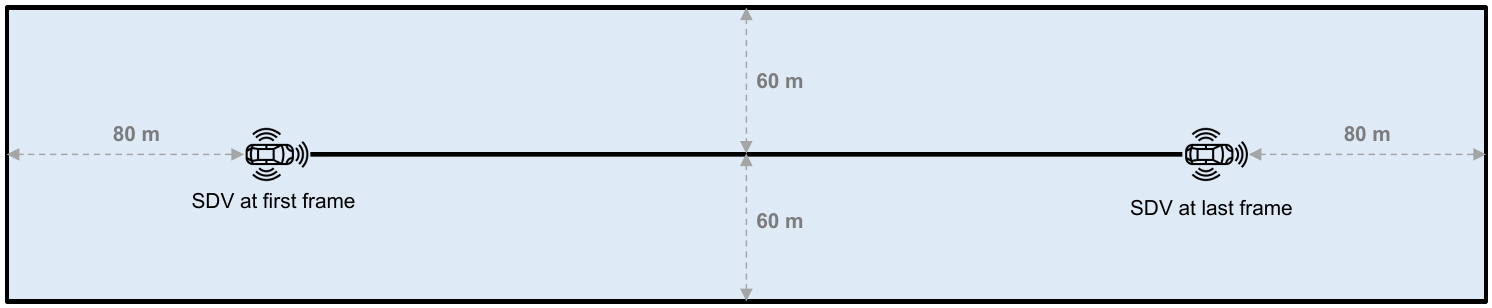}
     \caption{Region of interest of our scene representation.}
     \label{fig:supp_roi}
\end{figure*}

We begin by defining the region of interest for the static scene based on the trajectory of the self-driving vehicle (SDV).
The scene extends $80$ meters behind the SDV position at the first frame, and $80$ meters ahead of the SDV position at the last frame.
The scene width is $120$ meters and the scene height is $40$ meters.
Please see Figure~\ref{fig:supp_roi} for a birds-eye-view illustration of our region of interest.

Next, we generate an occupancy grid for the scene volume (see Section 3.2 in the main paper) and model it using multi-resolution feature grids.
For distant regions outside the volume, we employ a background model similar to~\cite{zhang2020nerf++,barron2022mip}.
For dynamic actors, each actor model is represented by an independent multi-resolution feature grids generated from a shared HyperNet.
Following~\cite{mueller2022instant}, we use a spatial hash function to map each feature grid to a fixed number of features.

\paragraph{Neural Feature Fields (NFFs) Details:}
To obtain the geometry $s$ and feature descriptor $\mathbf f$ from the \fg for both the static scene and actors (Eq. 1 in main paper), we employ two networks, denoted as $f_\text{bg}$ and $f_\mathcal{A}$.
Each of these networks consists of two sub-networks.
The first sub-network is an
MLP that takes the interpolated feature $\{\texttt{interp}(\mathbf x, \mathcal G^l)\}_{l=1}^L$ and predicts the geometry $s$ (signed distance value) and intermediate geometry feature, the second sub-network is an
MLP takes the intermediate geometry feature and viewpoint encoding as input and predicts the neural feature descriptor $\mathbf f$.
A convolutional neural network $g_\text{rgb}$ is applied on top of the rendered feature map and produces the final image.
We also have an additional decoder for lidar intensity ($g_\text{int}$).

\paragraph{Actor Model Details:}
Although we assume the dynamic actor tracklets are provided, they might be inaccurate, even when human-annotated, and lead to blurry results.
We find it advantageous to refine the actor tracklets during training.
For each dynamic actor $\mathcal A_i$ with trajectory initialized by a sequence of SE(3) poses $\{\mathbf{T}_{\mathcal A_i}^{t}\}_{t=1}^{T}$, we jointly optimize the rotation and translation components of the poses at each timestamp.
We parameterize the rotation component using a 6D representation~\cite{zhou2019continuity}.
We also incorporate a symmetry prior along the longitudinal axis for vehicle objects, which are the most common actors in self-driving scenes.
Specifically, we denote the query points and view direction in the canonical object coordinate (Front-Left-Up) as $\mathbf x_{\mathcal A_i}$ and $\mathbf d_{\mathcal A_i}$.
During training, we randomly flip the input point and view direction when querying the neural feature fields:
\begin{align}
\mathbf x'_{\mathcal A_i} =
\begin{pmatrix}
1 & 0 & 0 \\
0 & -1 & 0 \\
0 & 0 & 1
\end{pmatrix}
\mathbf x_{\mathcal A_i},\quad
\mathbf d'_{\mathcal A_i} =
\begin{pmatrix}
1 & 0 & 0 \\
0 & -1 & 0 \\
0 & 0 & 1 \\
\end{pmatrix}
\mathbf d_{\mathcal A_i}
\end{align}

\paragraph{Actor Behavior Model for Closed-loop Simulation:}
For simplicity, we modelled the counterfactual actor behaviors with heuristics.
While we only demonstrated human-specified trajectories in the paper, it is intuitive to to incorporate other models to control actors' behaviors, such as intelligent driver models \cite{idm, gidm} or deep-learning based models \cite{suo2021trafficsim, igl2022symphony}.

\paragraph{Run-time and Resources:}
Table~\ref{tab:profiling} reports training/inference time for \name and baselines on an A5000 GPU for a scenario with 31 actors (Figure 6 row 1).
We believe the run-time can be further optimized with recent advancements~\cite{chen2022mobilenerf, yu2021plenoctrees} and code optimization (\eg, customized CUDA kernels).
We can also balance run-time and realism by adjusting the image resolution, lidar rays sent, and the number of points sampled along the ray.

\begin{table}[h]
    \captionsetup{font=footnotesize}
    \centering
    \vspace{-5pt}
        \begin{tabular}{lccc}
            \toprule
             & Train (Hour) $\downarrow$ & Camera (FPS) $\uparrow$ & LiDAR (FPS) $\uparrow$ \\
            \midrule
            FVS[57] & $\approx$ 24  & 0.33 & - \\
            NSG[51] & $\approx$ 20 & 0.06 & - \\
            Instant-NGP[46] & 0.11 & 2.60 & - \\
            Ours & 1.67 & 1.25 & 11.76 \\
            \bottomrule
        \end{tabular}
    \vspace{-5pt}
    \caption{Comparison of training speed, inference speed for camera image (1080p) and LiDAR frame ($\approx 77k$ rays per frame).}
    \label{tab:profiling}
\end{table}

\section{Implementation Details for Baselines}
\label{sec:baseline_details}
\subsection{Free View Synthesis (FVS)}
\label{sec:fvs_details}

One key component of FVS is that it requires a proxy geometry to perform warping.
Instead of using COLMAP to perform structure-from-motion as in the original paper, we take advantage of the fact that the data collection platform records accurate LiDAR depth information for the scene.
For the static scene, we aggregate LiDAR points across all frames with dynamic points removed and then create triangle surfels for them.
We first estimate per-point normals from 200 nearest neighbors, and then we downsample the points into $4cm^3$ voxels~\cite{zhou2018open3d}.
After that, triangle surfels are created with a 5cm radius.
This includes static vehicles.
For dynamic objects, we create surfels from aggregated point clouds using human labeled 3D bounding boxes as correspondence.
To enhance the photorealism of the rendered images, we also add the same adversarial loss as in UniSim, which helps produce higher quality results.

To render a target image, a few source images need to be selected for warping.
During training, we randomly select from nearby frames (10 frames before and 10 frames after).
To test the interpolation during inference, we select the 2 most nearby frames.
To test the lane shift, we select every 4 frames starting from 12 frames before to 4 frames after current frame.
We heuristically tried multiple settings and found this setting usually achieves better balance between performance and efficiency.
Fewer source images will result in images with large unseen region, and the network may fail to inpaint it.
More source images may produce blurry results because of geometry misalignment and runs out-of-memory on the GPU.
The FVS is trained on 2 A5000 GPUs with learning rate $0.0001$, batch size 8 for $50,000$ iterations. Images are cropped to $256 \times 256$ patches during training.

\subsection{Instant-NGP}
Instant-NGP~\cite{mueller2022instant} achieves state-of-the-art NeRF rendering by introducing an efficient hash encoding, accelerated ray sampling and fully fused MLPs.
In our experiments, we scale the PandaSet scenes to fully occupy the unit cube and set \texttt{aabb\_scale} as 32 to handle the background regions (e.g., far-away buildings and sky) outside the unit cube.
We tuned \texttt{aabb\_scale} to be 1, 2, 4, 8, 16, 32, 64 and find 32 leads to the best performance.
The model is trained for 20K iterations and converges on the training views.
However, we find there are obvious artifacts including floating particles and large missing regions on the ground.
This is due to the radiance-geometry ambiguity~\cite{zhang2020nerf++} and limited viewpoints that are sparse and far apart in the driving scenarios compared to dense indoor scenes.
Therefore, to improve the geometry for better extrapolation (e.g., lane shift), we enhance the original method with LiDAR depth supervision.
Specifically, we aggregate the recorded LiDAR data and create a surfel triangle representation.
We follow the same surfel asset generation pipeline as stated in the implementation of FVS (see Sec.~\ref{sec:fvs_details}).
We then use the surfel mesh representation to render a depth image at each camera training viewpont.
We also tried to ignore the dynamic actors with rendered object masks but it leads to worse performance in terms of PSNR since all dynamic actors are removed.
We use the official repository~\footnote{\url{https://github.com/NVlabs/instant-ngp}} (commit id: \texttt{ca21fd399d0eec4d85fd31ac353116bc088f3f75}, Oct 19, 2022) for our experiments.

\subsection{Neural Scene Graph (NSG)}
We use the codebase on GitHub\footnote{\url{https://github.com/princeton-computational-imaging/neural-scene-graphs}} and adopt the default setting for the model and training parameters.
For data preprocessing, only dynamic vehicles are considered as object nodes, static vehicles are considered as part of the background.
Also, we only select five dynamic actors that are the most present in the log due to memory limitations.
We find with a larger number of actors the model runs out-of-memory during data loading on a 128GB memory machine because our image data is higher resolution ($1920\times1080$) than the KITTI data ($1242\times 375$).
NSG only works well on front-facing scenarios with only small movements along the camera viewing direction.
We tried to improve the background modeling by increasing the background points sampling and increasing model size.
However, we did not see significant improvement in the visual quality, while also making the computation cost much more expensive.
Therefore, we stick to the default setting.

\begin{figure}[t]
	\begin{center}
		\includegraphics[width=1.0\textwidth]{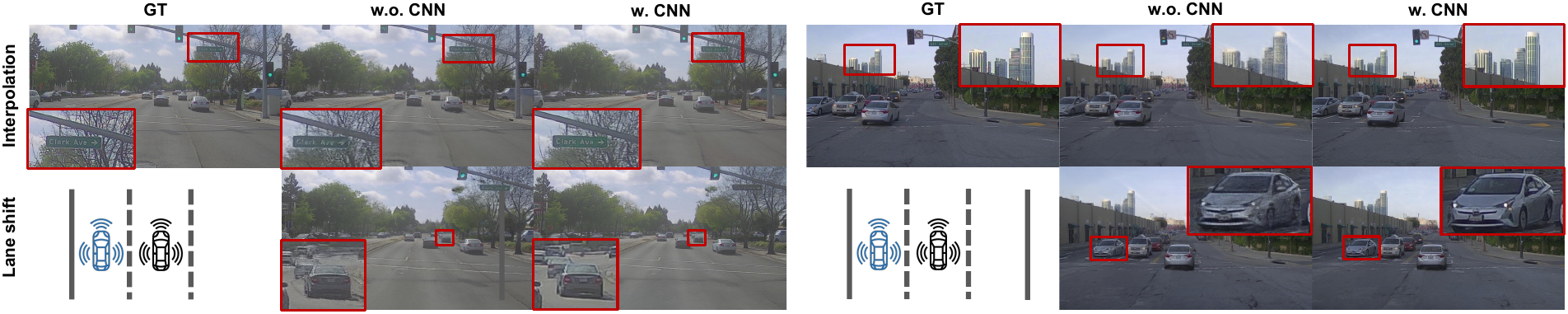}
	\end{center}
	\caption{\textbf{Effects of CNN Decoder}. The convolutional RGB decoder helps to make the rendered image more clean and learn more details.}
	\label{fig:cnn}
\end{figure}

\begin{figure}[t]
	\begin{center}
		\includegraphics[width=1.0\textwidth]{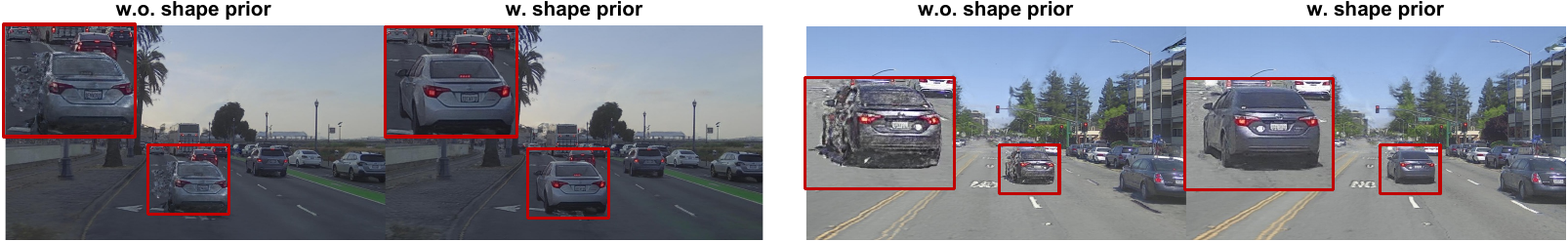}
	\end{center}
	\caption{\textbf{Effects of actor shape prior}. We show the unobserved side of the vehicle after SDV lane change, incorporating actor shape prior helps to reconstruct unobserved side of the vehicle.}
	\label{fig:actor}
\end{figure}

\subsection{LiDARsim}
We re-implement LiDARsim based on OptiX ray tracing engine~\cite{parker2010optix}.
We first create the asset bank following~\cite{manivasagam2020lidarsim}: Specifically, for background assets, we aggregate the LiDAR points across all the frames and remove all actors using 3D bounding box annotations.
For dynamic actors, we aggregate the LiDAR points inside the bounding boxes in the object-centric coordinate.
We then estimate per-point normals from 200 nearest neighbors with a radius of $20cm$ and orient the normals upwards for flat ground reconstruction.
We downsample the LiDAR points into 8cm voxels and create per-point triangle faces (radius $10cm$) according to the estimated normals.
Note that we also track the per-point intensity values during the above post-processing.
Given the asset bank, we place the background and all actors in its original locations and transform the scene to the LiDAR coordinate for ray-triangle intersections computation.
For LiDAR realism experiments, we created surfel assets using training frames ($0, 2, 4, \cdots, 78$) and performed raycasting using real LiDAR rays on held-out frames ($1, 3, 5, \cdots, 79$).
For perception evaluation and training, the assets assets are created and raycasted with all 80 frames.
Instead of using real LiDAR rays which are infeasible to obtain in real-world simulation, we set the sensor intrinsics (\eg, beam angle, azimuth resolution, etc) based on the public documents and raw LiDAR information~\footnote{\url{https://github.com/scaleapi/pandaset-devkit/issues/67}}.

\section{\name Experiment Details}
\label{sec:exp_details}

\subsection{Image Similarity Metrics}
Due to limited computation resources available to run NSG on all the logs in PandaSet, 10 scenes were selected for image similarity evaluation: \texttt{001,011,016,028,053,063,084,106,123,158}.
These scenes include a variety of both city and suburban areas, includes one night log (\texttt{063}), and one log containing no dynamic actors (\texttt{028}).
These 10 scenes are of a subset of the 24 selected for perception validation, described in the following section.

\subsection{Perception Evaluation and Training}
To study whether the sensor simulation is realistic with respect how autonomy perceives it, we first analyze the domain gap for perception tasks.
Specifically, we leveraged the state-of-the-art camera-based brids-eye-view (BEV) detection model BEVFormer~\cite{li2022bevformer} and evaluate on the full PandaSet.
Since there are no official train/val splits available, we split PandaSet into 79 and 24 sequences as training and validation sets by considering geographic locations, time span and day-night variety.
Specifically, we select the sequences \texttt{001,011,013,016,017,028,029,032,053,054,063,065,068,072,084,
086,090,103,106,110,112,115,123,158} as validation set and leave other 79 sequences as training.

We consider two setups (a) \textbf{Real2Sim}: evaluating the perception model trained on real data on simulated data; (b) \textbf{Sim2Real}: training perception models with simulated data data and testing on real data.
Specifically, we evaluate the real model on 24 simulated validation logs for Real2Sim and train perception models with 79 simulated training logs for Sim2Real.
For data augmentation experiments, we first use all simulation variations (log-replay, lane shift to the left for 0.5 and 2.0 meters) to train the detector (Sim).
Then we combine those simulation data with real data (Sim + Real) to retrain the detector.
Note that for all experiments, we train UniSim and other baselines with all image frames (\ie, frames 0 - 79).
As shown in Table 5 (main paper), incorporating UniSim lane shift variations help autonomy achieve superior detection performance compared to training with real data alone. More rigorous study on how to sufficiently leverage the simulated data (\eg, advanced actor behavior simulation, actor insertion, camera rotations, etc) is left as future works.

\paragraph{BEVFormer Implementation Details:}
We adapt the official repository~\footnote{\url{https://github.com/fundamentalvision/BEVFormer}} for the training and evaluation on PandaSet.
Specifically, we focus on single-frame vehicle detection with the front camera.
We ignore the actors that are out of the camera field-of-view during training and evaluation.
The models are trained in the vehicle frames with FLU convention ($x$: forward, $y$: left, $z$: up).
The region of interest is set as $x \in [0, 80m], y \in [-40m, 40m], z \in [-2m, 6m]$.
Due to memory limit, we adopt the BEVFormer-small architecture~\footnote{\url{https://github.com/fundamentalvision/BEVFormer/blob/master/projects/configs/bevformer/bevformer_small.py}} with a batch size of 2 per GPU.
We train the models using AdamW optimizer~\cite{loshchilov2017decoupled} for 20 epochs with the cosine learning rate schedule~\footnote{\url{https://pytorch.org/docs/stable/generated/torch.optim.lr_scheduler.CosineAnnealingLR.html}}.
Each model takes around 12 hours to train with 2$\times$ RTX A5000.
For data augmentation experiments, we generate the simulated data offline and train the models for 10 epochs as the datasets become three to four times larger compared to original PandaSet.

\begin{figure}[t]
	\begin{center}
		\includegraphics[width=1.0\textwidth]{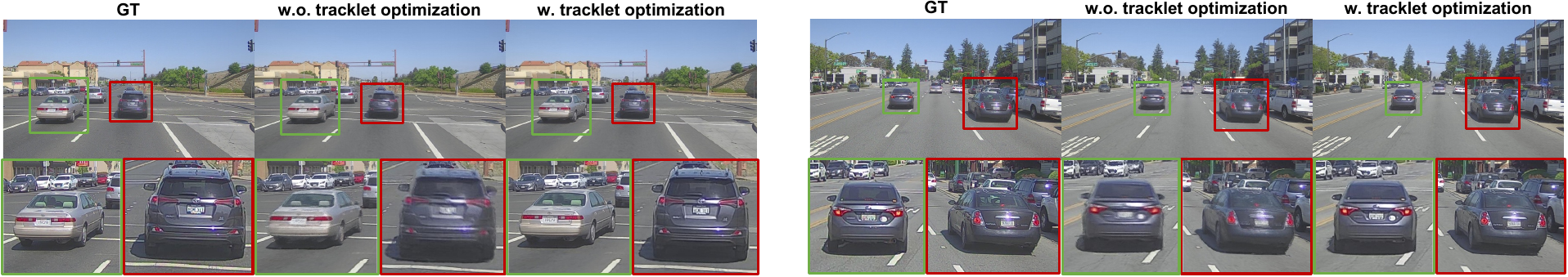}
	\end{center}
	\caption{\textbf{Effects of tracklet refinment}. Tracklet refinement helps to learn sharp details for dynamic actors.}
	\label{fig:tracklet_opt}
\end{figure}

\subsection{Open-loop Autonomy Evaluation}
\paragraph{Evaluation Set:}
In main-paper Sec. 4.5, we ran open-loop autonomy metrics on PandaSet scenes.
The prediction~\cite{cui2018multimodal} and planning modules~\cite{plt} require map information to operate, which PandaSet
does not have.
Maps were annotated for 9 scenes in PandaSet that are in the autonomy validation set.
These are a subset of the 10 scenes (excludes \texttt{063}) used for quantitative evaluation in main-paper Sec. 4.3 that also have LiDAR semantic segmentation from PandaSet to make map annotation possible.

\begin{figure}[t]
	\begin{center}
		\includegraphics[width=1.0\textwidth]{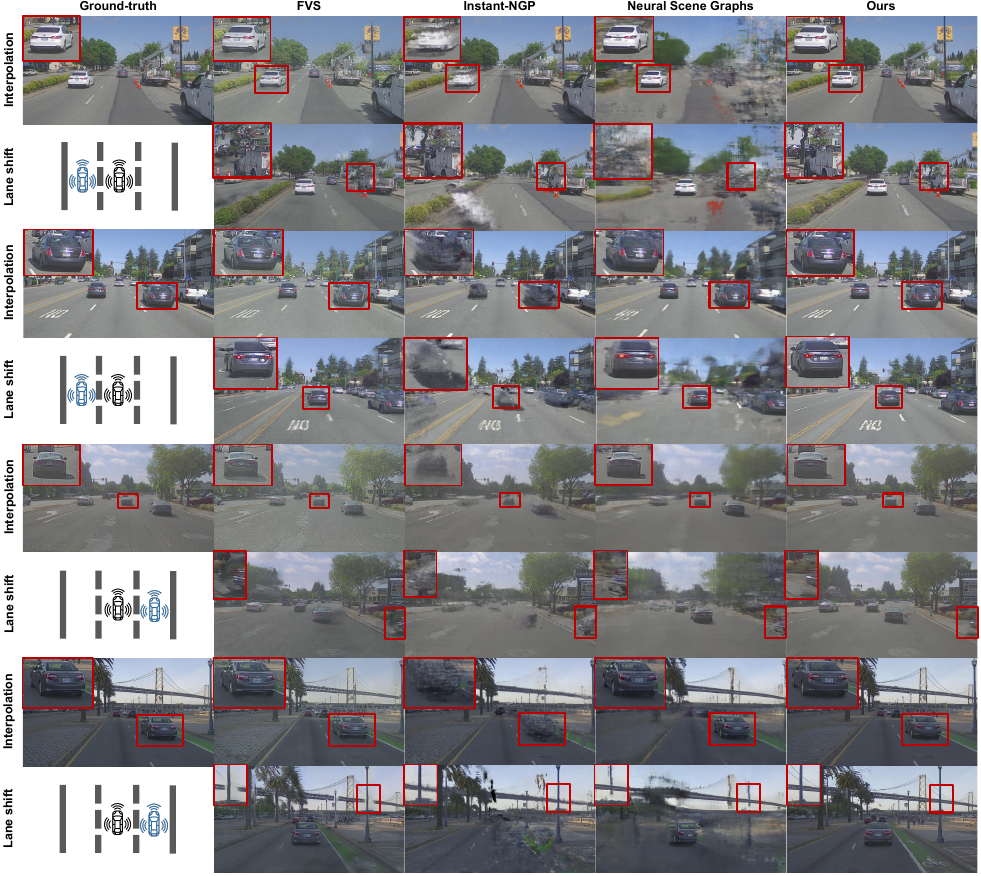}
	\end{center}
	\caption{\textbf{Qualitative comparison on more examples.} We show rendering results in both the interpolation and \textcolor{sdvcolor}{lane-shift} test settings.}
	\label{fig:qualitative}
\end{figure}

\begin{figure}[t]
	\begin{center}
		\includegraphics[width=1.0\textwidth]{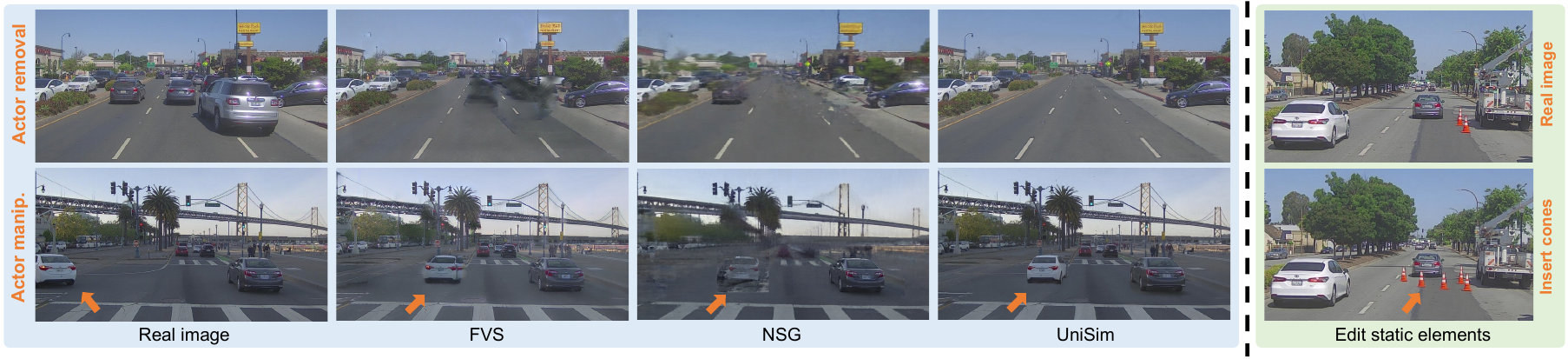}
	\end{center}
	\captionof{figure}{\small
    \textbf{Left:} Comparison to baselines on actor removal (top) and manipulation (bottom).
	\textbf{Right:} Edit static elements by copying cones.
	}
	\label{fig:manipulation}
\end{figure}

\begin{figure}[t]
	\begin{center}
		\includegraphics[width=1.0\textwidth]{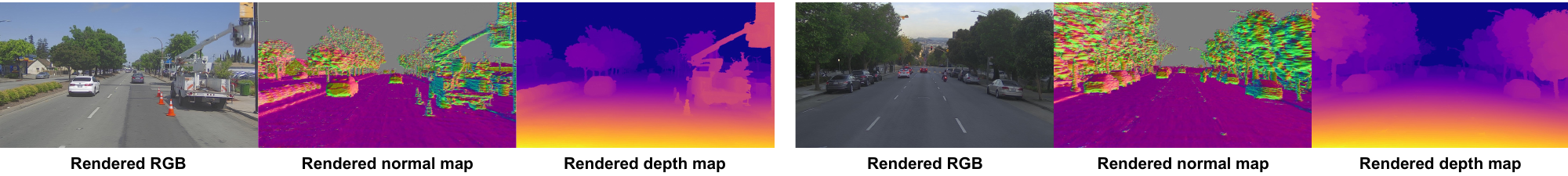}
	\end{center}
	\caption{\textbf{Rendered Scene Geometry}. For each example, from left to right: rendered RGB image, rendered normal map, and rendered depth map.}
	\label{fig:geometry}
\end{figure}

\begin{figure}[t]
	\begin{center}
	\includegraphics[width=1.0\textwidth]{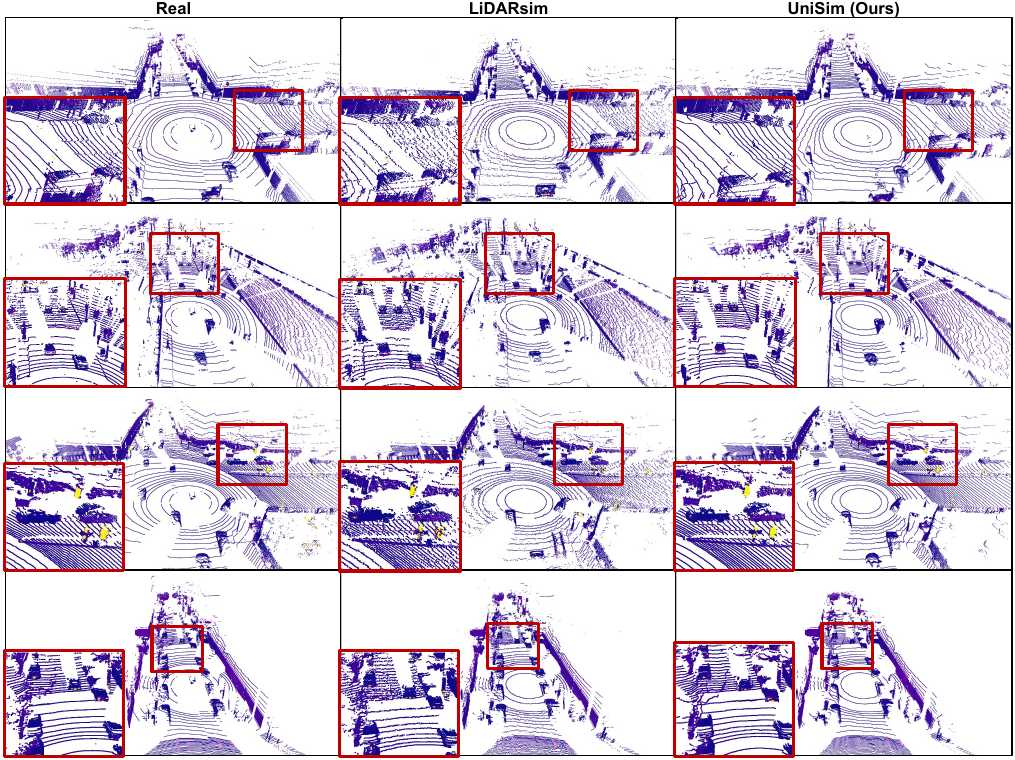}
	\end{center}
	\caption{\textbf{Comparison of LiDAR simulation on PandaSet.}
		Left: real lidar sweeps; Middle: simulated point clouds using LiDARsim~\cite{manivasagam2020lidarsim};
		Right: simulated point clouds using UniSim. Compared to LiDARsim, our neural rendering approach produces higher-fidelity LiDAR simulation with less noise and more continuous beam rings that are closer to real LiDAR scans.
		Note that the UniSim simulation results are more realistic with respect to how the autonomy perceives it (\ie, smaller Real2Sim and Sim2Real domain gap).
		}
\label{fig:lidarsim_comp}
\end{figure}

\paragraph{Open-loop Autonomy Metrics:}
We now describe in more detail the autonomy metrics computed for each module.

For detection, we compute a metric called detection agreement.
For detection agreement, we evaluate the real-trained BEVformer model on the real image and the simulated image, obtaining real-image and simulated-image
detection outputs.
We assign the real-image detections as the labels and compute the average precision (AP) between the simulated-image detections and the real-image detections (set as labels) to report the final result.
We compute the detection agreement at intersection-over-union (IoU) 0.3.
This metric measures whether the simulated sensor data generates the same true-positives and false-positives as
the real data at each frame.

For prediction, we compute the prediction average displacement error, where the motion forecasted trajectory is for a 6 second horizon at 10Hz frequency.
The prediction module selected for the autonomy is from Cui et. al.~\cite{cui2018multimodal}, which takes a rasterized map and 5 seconds past history tracks for each detected actor as input and outputs 6 second multi-modal trajectory horizons (number of modes = 6) for each actor.
The most-likely prediction mode for each actor generated in the real image by the prediction model is set as the label.
We then evaluate prediction outputs generated by simulated images by identifying matched actors between the label and predicted, and computing the minimum average displacement error (minADE) across the 6 predictions for that actor.
We report minADE at the maximum F1 score for each simulation method, where a true positive is defined as a matched detected actor from the simulated sensor data and real sensor data with IoU=0.3.

For planning, we compute the plan consistency metric. At each timestep, the autonomy model outputs a 5 second plan it would execute given the prediction output and map information.
We compute the final displacement error between the 5 second end point when running on real images and when running on simulated images at each timestep.

\section{Additional Experiments and Analysis}
\label{sec:additional_results}

\subsection{Ablation Study}
\paragraph{Convolutional RGB Decoder:}
The Convolutional RGB decoder improves the overall image quality by spatial relation reasoning and increases model capacity.
Figure~\ref{fig:cnn} shows a visual comparison on interpolation and lane shift settings.
It can be seen from the figure that the convolutional RGB decoder helps to make the rendered image more clean and perserve more details.

\paragraph{Actor Shape Prior:}
We incorporate hypernetwork and symmetry prior to learn better assets.
Figure~\ref{fig:actor} shows that incorporating these actor priors helps to reconstruct unobserved regions of the vehicle when SDV performs a lane shift.

\paragraph{Tracklet Refinement:}
We show a visual comparison training a model with and without tracklets optimization in Figure~\ref{fig:tracklet_opt}.
Optimizing tracklets improves reconstruction of the dynamic actors and makes sharpens their detail.

\begin{figure}[t]
	\begin{center}
		\includegraphics[width=1.0\textwidth]{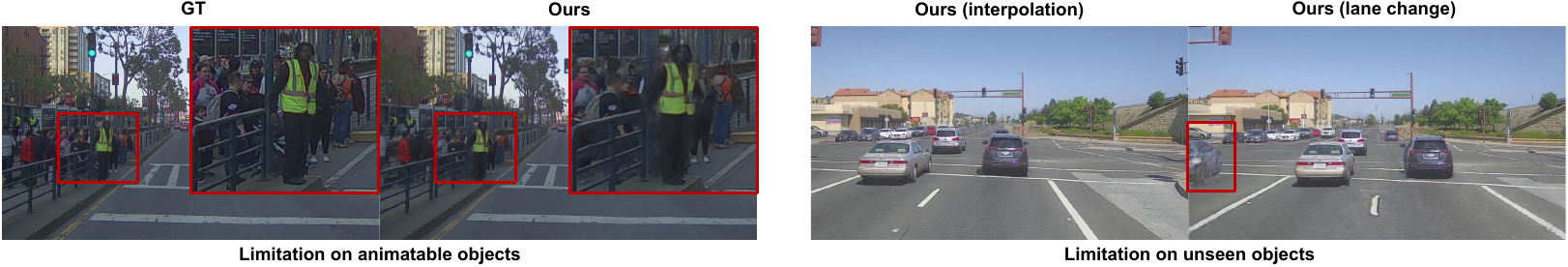}
	\end{center}
	\caption{\textbf{Limitations of \name.}
	Left: Our method learns blurry results for animatable objects since we do not model the animation.
	Right: Our method has difficult rendering completely unseen objects after performing SDV lane change.
	}
	\label{fig:limitations_qual}
\end{figure}

\subsection{Additional Camera Simulation Results}
\paragraph{Realism Evaluation:}
Please see Figure~\ref{fig:qualitative} for more examples on qualitative comparisons with the baseline models.
We show rendered results in both the interpolation and lane-shift test settings.
We also show the comparison to baselines on actor removal and manipulation in Figure~\ref{fig:manipulation} left.
Additionally, we visualize the rendered normal (normalized gradient of SDF $s$) and depth map in Figure~\ref{fig:geometry}.

\paragraph{Editing Scene Elements:}
Similar to dynamic actors, we can represent and edit static elements using bounding box annotations.
Figure~\ref{fig:manipulation} right shows we copy the cones to new placements and create new scenarios.

\subsection{Additional LiDAR Simulation Results}
To investigate the realism of LiDAR simulation, we compare UniSim with the state-of-the-art approach LiDARsim~\cite{manivasagam2020lidarsim} in the \textit{log-replay} setting.
Figure~\ref{fig:lidarsim_comp} show qualitative examples of real LiDAR scans (left), and simulated LiDAR scans using LiDARsim and UniSim.
LiDARsim conducts raycasting with the aggregated surfel meshes and produces incomplete beam rings and noisy scans.
This is because the surfel meshes created with the original LiDAR points can have noisy normal estimates and incomplete shape.
Compared to LiDARsim, the LiDAR sweeps simulated by UniSim is more accurate and smoother.
This is because our neural representations capture the underlying geometry better.
Moreover, UniSim can simulate the intensity values more smoothly and accurately (See the third row in Figure~\ref{fig:lidarsim_comp}).

\begin{table}[]
	\centering
	\begin{tabular}{lccccc}
		\toprule
		\textbf{Real2Sim} (\textit{replay}) & \textbf{mAP} & \textbf{AP@0.1} & \textbf{AP@0.3} & \textbf{AP@0.5} & \textbf{AP@0.7} \\
		\midrule
		FVS  & 37.2 & 63.4 & 46.7 & 27.1 & 11.5 \\
		Instant-NGP & 22.6 & 40.9 & 28.8 & 15.4 & ~~5.4 \\
		UniSim & \textbf{40.2} & \textbf{67.2} & \textbf{50.7} & \textbf{30.8} & \textbf{12.2} \\
		\midrule
		Real & 40.9 & 68.1 & 51.4 & 31.3 & 12.7 \\
		\bottomrule
	\end{tabular}
	\caption{Real2Sim domain gap in \textit{replay} setting.}
	\label{tab:real2sim_replay}
\end{table}

\begin{table}[]
	\centering
	\begin{tabular}{lccccc}
		\toprule
		\textbf{Real2Sim} (\textit{lane-shift}) & \textbf{mAP} & \textbf{AP@0.1} & \textbf{AP@0.3} & \textbf{AP@0.5} & \textbf{AP@0.7} \\
		\midrule
		FVS & 30.3 & 54.1 & 38.9 & 21.7 & 6.6 \\
		Instant-NGP & 18.1 & 34.3 & 22.2 & 11.9 & 3.8 \\
		UniSim & \textbf{37.0} & \textbf{63.6} & \textbf{47.8} & \textbf{27.2} & \textbf{9.4} \\
		\bottomrule
	\end{tabular}
	\caption{Real2sim domain gap in \textit{lane-shift} setting.}
	\label{tab:real2sim_shift}
\end{table}

\begin{table}[]
	\centering
	\begin{tabular}{lccccc}
		\toprule
		\textbf{Sim2Real} (\textit{replay}) & \textbf{mAP} & \textbf{AP@0.1} & \textbf{AP@0.3} & \textbf{AP@0.5} & \textbf{AP@0.7} \\
		\midrule
		FVS & 38.7 & 66.2 & 48.8 & 28.7 & 11.1 \\
		Instant-NGP & 34.0 & 62.3 & 43.9 & 22.9 & ~~6.7 \\
		UniSim & \textbf{39.9} & \textbf{67.6} & \textbf{51.2} & \textbf{29.4} & \textbf{11.4} \\
		\midrule
		Real & 40.9 & 68.1 & 51.4 & 31.3 & 12.7 \\
		\bottomrule
	\end{tabular}
	\caption{Sim2Real domain gap in \textit{replay} setting.}
	\label{tab:sim2real_replay}
\end{table}

\begin{table}[]
	\centering
	\begin{tabular}{lccccc}
		\toprule
		\textbf{Sim2Real} (\textit{lane-shift}) & \textbf{mAP} & \textbf{AP@0.1} & \textbf{AP@0.3} & \textbf{AP@0.5} & \textbf{AP@0.7} \\
		\midrule
		FVS & 32.2 & 59.9 & 41.5 & 21.0 & 6.5 \\
		Instant-NGP & 26.5 & 54.3 & 34.8 & 13.9 & 2.9 \\
		UniSim & \textbf{37.1} & \textbf{64.9 } & \textbf{47.5} & \textbf{26.4} & \textbf{9.5} \\
		\bottomrule
	\end{tabular}
	\caption{Sim2Real detection performance in \textit{lane-shift} setting.}
	\label{tab:sim2real_shift}
\end{table}

\begin{table}[]
	\centering
	\begin{tabular}{lccccc}
		\toprule
		\textbf{Data Augmentation} & \textbf{mAP} & \textbf{AP@0.1} & \textbf{AP@0.3} & \textbf{AP@0.5} & \textbf{AP@0.7} \\
		\midrule
		Real & 40.9 & 68.1 & 51.4 & 31.3 & 12.7 \\
		\midrule
		Sim (FVS) & 39.2 & 68.0 & 50.1 & 28.5 & 10.1 \\
		Sim (Instant-NGP) & 32.4 & 60.7 & 41.1 & 21.3 & ~~6.6 \\
		Sim (UniSim) & \textbf{41.4} & \textbf{71.4} & \textbf{54.7} & \textbf{29.5} & \textbf{10.1} \\
		\midrule
		Real + Sim (FVS) & 41.1 & 71.6 & 53.6 & 29.6 & ~~9.6 \\
		Real + Sim (Instant-NGP) & 40.1 & 70.1 & 51.5 & 29.0 & ~~9.9 \\
		Real + Sim (UniSim) & \textbf{42.9} & \textbf{71.9} & \textbf{54.3} & \textbf{32.7} & \textbf{12.8} \\
		\bottomrule
	\end{tabular}
	\caption{\textbf{Augmenting real data with simulation.}
		Using UniSim data only to train the perception model is even better than training with all real data.
		UniSim augmentation yields a significant performance gain.
		In contrast, baseline data augmentation brings marginal gain or harms performance
	}
	\label{tab:data_aug_full}
\end{table}

\subsection{Additional Perception Evaluation and Training for Camera Simulation}
We report detailed detection metrics for perception evaluation and training for better reference.
Specifically, we report the average precision (AP) at different IoU thresholds: 0.1, 0.3, 0.5, 0.7.
The mean average precision (mAP) is calculated by $\mathrm{mAP = (AP@0.1 + AP@0.3 + AP@0.5 + AP@0.7) / 4.0}$.
The Real2Sim results are shown in Table~\ref{tab:real2sim_replay} (replay) and Table~\ref{tab:real2sim_shift} (lane-shift).
The Sim2Real results are shown in Table~\ref{tab:sim2real_replay} (replay) and Table~\ref{tab:sim2real_shift} (lane-shift).
The data augmentation experiments are presented in Table~\ref{tab:data_aug_full}.

\subsection{Perception Evaluation and Training for LiDAR Simulation}
Besides camera simulation, we also analyze if the simulated LiDAR reduces the domain gap for perception tasks compared to an
existing SoTA LiDAR simulator LiDARsim~\cite{manivasagam2020lidarsim}.
We leveraged the LiDAR-based birds-eye-view detection model PIXOR~\cite{yang2018pixor} and tested Real2Sim and Sim2Real setups.
We used the same training validation splits as the camera perception model and considered the replay setting.
Specifically, we use all frames to train the models (creating assets for LiDARsim) and test with our approximated sensor configurations.
We focus on multi-sweep (5 frame) vehicle detection with a region of interest (ROI) is set as $x \in [0, 80m], y \in [-40m, 40m], z \in [-2m, 6m]$.
We report the average precision (AP) at different IoU thresholds: 0.1, 0.3, 0.5, 0.7. 0.8.
The mean average precision (mAP) is calculated by $\mathrm{mAP = (AP@0.1 + AP@0.3 + AP@0.5 + AP@0.7 + AP@0.8) / 5.0}$.

As shown in Table~\ref{tab:real2sim_lidar} and Table~\ref{tab:sim2real_lidar},
UniSim results in smaller domain gap in both Real2Sim and Sim2Real compared to LiDARsim especially when the IoU threshold is strict (e.g. AP@0.7, AP@0.8).
This indicates our approach further bridges the gap between simulation and real world and can help better evaluate and train autonomy.
We have performed two analyses on perception model performance on camera and LiDAR sensors separately with the same UniSim models generating both synthetic data.
Future investigation would include performing multi-sensor perception experiments to further understand the domain gap of UniSim.

\begin{table}[]
	\centering
	\begin{tabular}{lcccccc}
		\toprule
		\textbf{Real2Sim} & \textbf{mAP} & \textbf{AP@0.1} & \textbf{AP@0.3} & \textbf{AP@0.5} & \textbf{AP@0.7} & \textbf{AP@0.8} \\
		\midrule
		LiDARsim & 74.9 & 88.3 & 86.3 & 82.7 & 69.1 & 48.3 \\
		UniSim & \textbf{77.0} & \textbf{90.1} & \textbf{88.4} & \textbf{84.4} & \textbf{71.5} & \textbf{50.8} \\
		\midrule
		Real & 80.8 & 93.6 & 92.0 & 88.0 & 75.2 & 55.2 \\
		\bottomrule
	\end{tabular}
	\caption{Real2Sim detection performance for LiDAR simulation.}
	\label{tab:real2sim_lidar}
\end{table}

\begin{table}[]
	\centering
	\begin{tabular}{lcccccc}
		\toprule
		\textbf{Sim2Real} & \textbf{mAP} & \textbf{AP@0.1} & \textbf{AP@0.3} & \textbf{AP@0.5} & \textbf{AP@0.7} & \textbf{AP@0.8} \\
		\midrule
		LIDARsim & 75.9 & 91.6 & \textbf{89.9} & \textbf{85.6} & 70.6 & 42.0 \\
		UniSim & \textbf{77.9} & \textbf{92.0} & 89.8 & 85.4 & \textbf{72.0} & \textbf{50.3} \\
		\midrule
		Real & 80.8 & 93.6 & 92.0 & 88.0 & 75.2 & 55.2 \\
		\bottomrule
	\end{tabular}
	\caption{Sim2Real detection performance for LiDAR simulation.}
	\label{tab:sim2real_lidar}
\end{table}

\subsection{Other Explorations}
We have observed that certain commonly-used techniques in novel-view-synthesis did not offer significant improvements with our architecture and data setting on our initial attempts.
Specifically, we found that hierarchical or fine-grained sampling did not significantly enhance realism but resulted in a considerable decrease in processing speed, possibly due to the majority of the scene being distant from the self-driving vehicle.
Additionally, optimizing exposure per frame did not improve realism as the exposure does not change within a single log snippet for a single camera.
We also attempted multi-camera training, but found that several of the side cameras in PandaSet had mis-aligned calibration and slightly different exposures.
Optimizing the camera poses and exposure in this setting did not offer improvements, potentially due to limited data from a single 8 second log.
Additionally, more training iterations and deeper architecture did not provide significant improvements, but rather increased the complexity of the model.
\section{Limitations and Future Works}
\label{sec:limitation}

\name has several limitations.
Our neural \fg cannot render lighting variations, as the modeled radiance bakes the lighting into the representation.
Future work involves explicitly modelling and manipulating scene lighting~\cite{srinivasan2021nerv, boss2021neural, zhang2021nerfactor} and weather.
We also do not model animation (\eg, turn signals, traffic lights) or actor deformation (\eg walking) and hope to incorporate ideas from works~\cite{park2020deformable, xu2022deforming, yuan2022nerf, peng2021neural} that tackle this.
We also have artifacts when rendering areas that were previously outside the field-of-view.
Figure~\ref{fig:limitations_qual} shows qualitative visualizations of these phenomena.
We hope future work in this direction can further enhance sensor simulation realism.

\end{document}